\documentclass{article}
\usepackage{microtype}
\usepackage{hyperref}
\PassOptionsToPackage{table}{xcolor}
\expandafter\def\csname ver@algorithmic.sty\endcsname{}
\usepackage[preprint]{icml2026}
\newif\ifanon

\usepackage{amsmath,amsfonts,amssymb,bm}
\usepackage{amsthm}
\usepackage{algorithmicx}
\usepackage{algpseudocode}
\usepackage{graphicx}
\usepackage{subcaption}
\usepackage{textcomp}
\usepackage{booktabs}
\usepackage{makecell}
\usepackage{tablefootnote}
\usepackage{multirow}
\usepackage{enumitem}
\usepackage{mathtools}
\usepackage{tikz}

\theoremstyle{plain}
\newtheorem{theorem}{Theorem}
\newtheorem{proposition}{Proposition}
\newtheorem{lemma}{Lemma}

\newtheorem{assumption}{Assumption}
\theoremstyle{definition}
\newtheorem{definition}{Definition}
\theoremstyle{remark}
\newtheorem{remark}{Remark}

\newenvironment{widefigure}[1][tp]{\begin{figure*}[#1]}{\end{figure*}}
\newenvironment{widetable}[1][tp]{\begin{table*}[#1]}{\end{table*}}
\newenvironment{spanfigure}[1][tbp]{\begin{figure}[#1]}{\end{figure}}
\newcommand{\figthreewidth}{\linewidth}
\newcommand{\figfourwidth}{\linewidth}
\newenvironment{spantable}[1][tp]{\begin{table*}[#1]}{\end{table*}}

\newcommand{\figAonesub}{0.31\columnwidth}
\newcommand{\figAonebreak}{\hfill}%
\newcommand{\figAtwowidth}{0.45\textwidth}
\newcommand{\figAthreewidth}{0.75\columnwidth}

\newcommand{\tableonestdnote}{}
\newcommand{\proofsketch}{Proof sketch}
\newcommand{\tcite}[2][]{\citet{#2}}

\newenvironment{credits}{\renewcommand{\subsubsection}[1]{\section*{##1}}}{}
\providecommand{\ackname}{Acknowledgements}
\providecommand{\Description}[1]{}

\newcommand{\norm}[1]{\left\lVert#1\right\rVert}
\newcommand{\nodot}{\spacefactor=3000{}}
\definecolor{linkblue}{rgb}{0,0.08,0.45}%
\hypersetup{colorlinks=true,linkcolor=linkblue,citecolor=linkblue,urlcolor=linkblue,filecolor=linkblue}

\anonfalse

\icmltitlerunning{STORMGUIDE-FL: Stochastic Remote Guidance for Federated Streaming Data}

\begin{document}

\twocolumn[
  \icmltitle{STORMGUIDE-FL: Stochastic Remote Guidance for\\
    Efficient Mining of Federated Streaming Data}

  \begin{icmlauthorlist}
    \icmlauthor{Manuel R{\"o}der}{thws}
    \icmlauthor{Frank-Michael Schleif}{thws}
  \end{icmlauthorlist}

  \icmlaffiliation{thws}{Technical University of Applied Sciences W{\"u}rzburg-Schweinfurt, W{\"u}rzburg, Germany}
  \icmlcorrespondingauthor{Manuel R{\"o}der}{manuel.roeder@thws.de}

  \icmlkeywords{Federated Learning, Streaming Data, Distributed Data Mining, Evolving Environments}

  \vskip 0.3in
]

\printAffiliationsAndNotice{}

\begin{abstract}
Federated Learning enables collaborative model training across distributed edge devices (“clients”) while preserving data privacy.
However, real-world deployments involve clients that observe continuous data streams from evolving environments while operating under strict resource limitations.
Existing solutions still face two major challenges: local concept drift, where each client’s data distribution shifts over time, and global client drift, arising from asynchronous engagement and imbalanced contributions across heterogeneous training participants, %
formally described as dual client heterogeneity.
We propose \textsc{STORMGUIDE-FL}, a federated optimization and data-mining framework that mitigates both types of drift using a server-guided pipeline for data acquisition and local training.
For data acquisition, a lightweight client-side sampling strategy incrementally selects informative and diverse instances from non-stationary, non-IID local streams under bounded memory, compute, and labeling budget.
For local training, the server's remote guidance is embedded directly into local optimization as an update-level correction,
counteracting client drift without discarding scarce labeled samples.
We empirically verify the effectiveness of the proposed sampling strategy on federated data streams across gradual, sudden, and recurring concept drift, and demonstrate that its gains compound with the update-level drift correction.
Experimental results further show improved speed and robustness of global model convergence under resource-constrained, heterogeneous conditions.

\end{abstract}

\section{Introduction}
The rapid evolution of digital technologies and the exponential growth of data, in conjunction with the increasing demand for data protection by society~\citep{EuropeanParliament2016a,adppa2022}, have underscored the significant importance of \emph{Federated Learning} (\textbf{FL})~\citep{mcmahan2017} in the modern era of artificial intelligence.
This innovative methodology embodies the shift towards decentralized data processing that enables collaborative learning without direct data exchange:
advanced FL systems harness multiple nodes, diversified access points, and distributed collection mechanisms, organized into a server-client meta framework instead of relying on centralized repositories.
Although the acquisition process of data points employed to train the global FL model allows a high degree of flexibility, this also introduces various challenges due to the non-stationarity of the data observations and the independence of the FL clients in real-world scenarios.

To address these shortcomings, we introduce a novel, data-centric approach where the FL server remotely guides client-side sampling from evolving data streams.
Instead of treating data acquisition and drift correction as separate concerns, our method lets the server's guidance signal act on both: it shapes which samples are mined from the stream and steers the local parameter updates computed on them toward the global optimization objective.
\begin{figure}[t!]
    \centering
    \includegraphics[width=1.0\columnwidth]{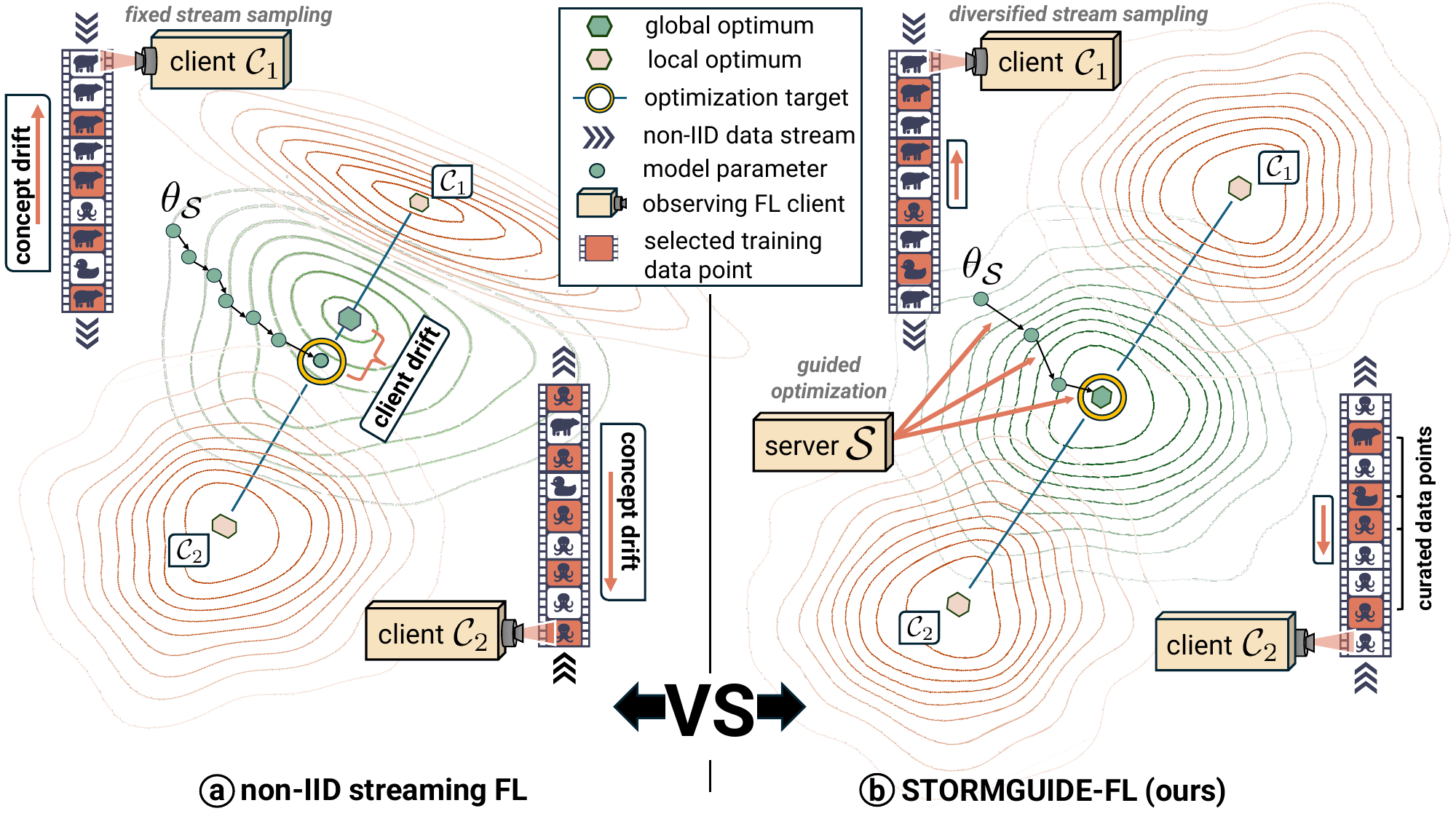}
    \caption{Loss landscapes w.r.t. model parameters $\theta_{\mathcal{S}}$ and local data of two FL clients $\mathcal{C}_1$ and $\mathcal{C}_2$, both observing separate, non-IID data streams (e.g., two trail cameras monitoring diverse wildlife under changing environments):
    \textcircled{a} The baseline for non-IID streaming FL, where uniform stream sampling and weight averaging lead to \emph{client drift} (optimization target $\neq$ global optimum);%
    \, \textcircled{b} \textsc{STORMGUIDE-FL} locally mitigates the impact of dynamic data observations in FL training through diversified stream sampling and server-guided optimization, rendering the refinement of the distributed model parameters towards the global optimum more focused and robust.}    
    \label{fig:motivation}
    \Description{The figure consists of two side-by-side panels, labeled (a) and (b), illustrating how federated learning clients behave under changing data conditions, known as concept drift. Each panel shows two clients, C$_1$ and C$_2$, represented by gray boxes containing small icons symbolizing data streams. Behind them are contour maps indicating the optimization landscape, where green and white circles mark global and local optima, respectively. In panel (a), client C$_1$’s optimization path drifts away from the shared global optimum due to differing, non-IID data distributions, shown by a red arrow labeled “client drift.” In panel (b), both clients’ optimization paths align more closely, converging toward the global optimum $\theta_\mathcal{S}$, illustrating improved adaptation and stability under concept drift. A legend in the center explains the symbols, including model parameters, optimization targets, and data points, emphasizing the relationship between client behavior and model convergence in a federated learning environment.}
\end{figure}

\subsection{Background and Motivation}
\label{subsec:intro_background}
Client edge devices deployed in production environments such as trail cameras~\cite{preetha2025fedwildlife}, medical sensors~\cite{feng2024flhealthcare}, financial monitoring systems~\cite{ardakani2023fedfinance} and industrial equipment~\cite{Jisu2023maintenance}, continuously observe \emph{non identically and independently distributed} (\textbf{non-IID}) data streams. In this setup, processing, storing, and communicating becomes prohibitively expensive, particularly for clients with limited resource availability.
Unlike traditional static datasets, these continuous streams of sensed observations arrive sequentially, typically without labels, and are subject to various forms of \emph{concept drift}~\citep{hinder2024} over time.
Statistical properties of target variables can shift abruptly, such as when new regulations alter financial reporting requirements overnight, or gradually, as patient demographics change in a hospital over months or years~\citep{Chibwe2024_hospital}.
Such changes give rise to concept drift in the local scope, where each FL client must continually adapt to evolving local data distributions in order to remain accurate and relevant.
In contrast, the global FL learning objective is confronted with its own challenges:
standard FL algorithms, such as \textsc{FedAvg}~\citep{mcmahan2017}, assume static or mildly heterogeneous clients.
However, in real-world implementations involving sequential non-IID data and naive sampling strategies that are not tailored to the specific characteristics of the data, \emph{client drift}~\citep{shi2022optimization} emerges.
This phenomenon refers to persistent divergence in client updates due to asynchronous or skewed observation histories, which leads to poor convergence and degraded global model performance~\citep{pmlr-v119-karimireddy20a}.

In combination, these effects motivate the formulation of the \emph{dual client heterogeneity} problem: local non-stationarity due to concept drift and global inconsistency due to client drift.
The illustration in \figurename~\ref{fig:motivation}~\textcircled{a} demonstrates intuitively how dual drift impacts the performance of global FL model optimization and outlines the mitigation effects of our approach:
By strategically identifying and extracting the most valuable observations from local client data streams rather than relying on arbitrary sampling in the first step, our solution systematically uncovers relevant data points for global model improvement and adaptation to changing conditions.
To further steer the FL process for global learning effectively, our approach integrates \emph{Remote-Guided Local Model Updates} in the second step, where a server remote guidance variate corrects every local training step, keeping the updates computed on the diversified set of data samples consistent with the global optimization trajectory.%

The overarching objective of this data mining pipeline is to identify the most valuable observations of each local stream and to integrate them into FL training under continuous global guidance.
In combination, both mechanisms align the clients' local optimization trajectories w.r.t. model parameters, thereby enabling a more robust, global model optimization process, even under the pressure of dual client heterogeneity (see \figurename~\ref{fig:motivation}~\textcircled{b}).

\begin{widefigure}[tb]
    \centering
    \includegraphics[width=\textwidth]{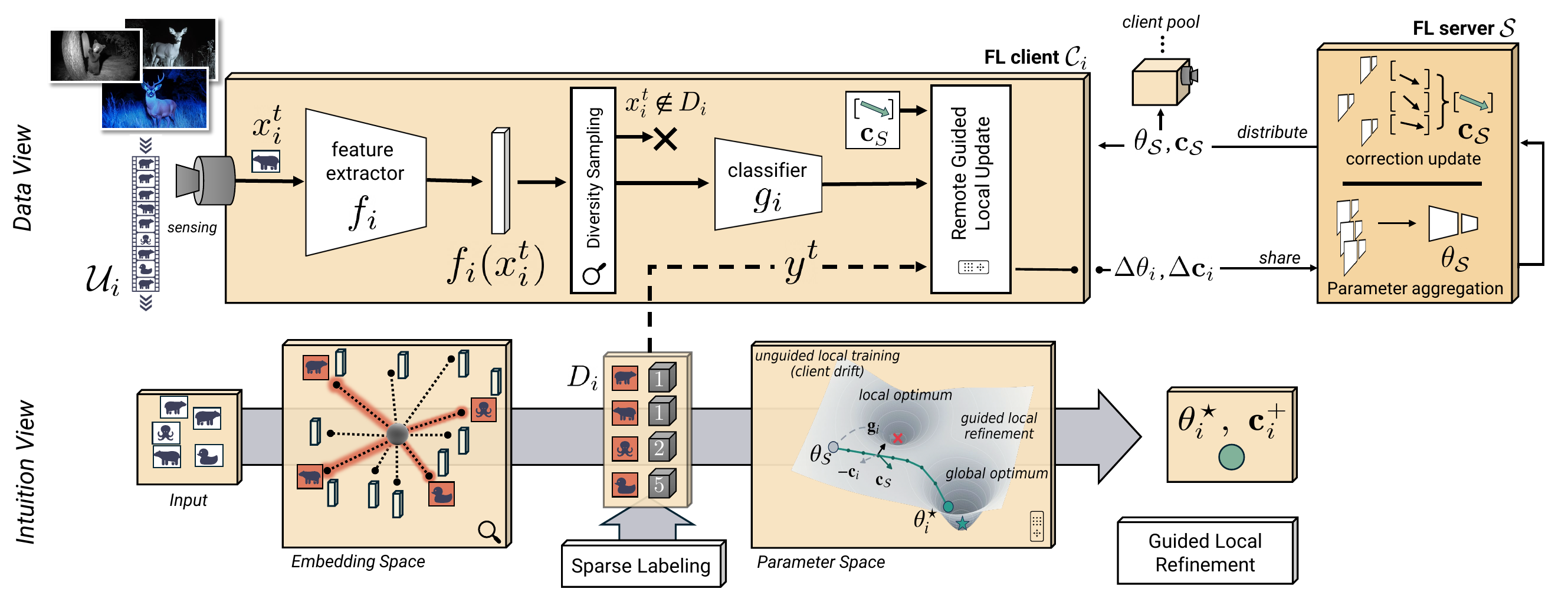}
    \caption{Overview of \textsc{STORMGUIDE-FL} from two perspectives, \textit{Data View} and \textit{Intuition View}, in an exemplary wildlife monitoring scenario.
    Both follow data point $x^t_i$, sensed from stream $\mathcal{U}_i$ on FL client $\mathcal{C}_i$ at timestep $t$.
    \textit{One-Pass Diversity Sampling} (OPDS-FL, Section~\ref{subsec:methodology_stage_1}) assesses sample informativeness based solely on the feature geometry of the representation $f_i(x^t_i)$.
    Accepted and annotated pairs $(x^t_i, y^t_i)$ form the local training set $D_i$ for the \emph{Remote-Guided Local Model Update} (Section~\ref{subsec:methodology_stage_2}), where the server remote guidance variate $\mathbf{c}_{\mathcal{S}}$ corrects every gradient step toward the global optimization trajectory.
    The resulting guided refinement $\theta^{\star}_i$ is folded into the updated client variate $\mathbf{c}^{+}_i$, and the deltas $(\Delta\theta_i, \Delta\mathbf{c}_i)$ are sent back to the FL server $\mathcal{S}$ for aggregation (Section~\ref{subsec:methodology_server_control}).
    }
    \label{fig:main_overview}
    \Description{Schematic overview of the federated learning framework in two sections. The upper Data View shows a client sensing stream data, extracting an embedding, and passing it to a one-pass diversity sampling module that accepts or rejects each sample under a labeling budget. Accepted samples are annotated and collected into the local training set, and a remote-guided local update corrects every training step with a server-provided guidance variate. The server above the client aggregates the returned parameter and variate deltas and redistributes global parameters and guidance. The lower Intuition View shows three stages: diverse picks in the embedding space, sparse labeling, and a parameter space in which the unguided trajectory drifts toward a client-specific local optimum while the guided trajectory converges toward the global optimum.}
\end{widefigure}

\subsection{Main Contributions}
\label{subsec:intro_contributions}
In this work, we introduce \textsc{STORMGUIDE-FL}, a novel FL algorithmic framework designed for learning from distributed streaming data under non-stationary conditions and resource limitations. 
Our key contributions are summarized below:
\begin{itemize}
    \item \textbf{SOTA Framework Extension:}
       We extend FL-Bench~\citep{tanFLBench} to support sampling and training on non-IID streaming data at scale.
       This enhancement enables realistic benchmarking of FL methods in evolving environments and provides a standardized testbed for our experiments in Section~\ref{sec:experiments} and a convenient entry point for future research.
     \item \textbf{Problem Formalization:} We present a novel formal definition of \emph{dual client heterogeneity} (Definition~\ref{def:dual_heterogeneity}) that separates inter-client skew from intra-client drift and anchors both our experimental design and convergence analysis.
     \item \textbf{Remote-Guided Learning Pipeline:} We introduce a server-orchestrated client pipeline that couples one-pass diversity sampling from non-IID streams under strict resource and labeling constraints (Section~\ref{subsec:methodology_stage_1}) with guidance-corrected local updates on the mined samples (Section~\ref{subsec:methodology_stage_2}), mitigating both forms of drift at their origin.
\end{itemize}

\section{Related Work}
\label{sec:relatedwork}
In the following, we review four research directions central to our approach:
\emph{non-IID FL}, addressing inter-client skew, and \emph{concept drift detection} on edge devices, addressing intra-client drift.
We conclude this section with their intersection, \emph{FL for data streams} and \emph{federated continual learning}, which inherits the complexities of both fields.

\paragraph{Non-IID FL and concept drift detection\nodot} have largely been studied separately.
\textsc{FedProx}~\citep{li2020} improves convergence under client heterogeneity by adding a proximal term to local objectives, \textsc{SCAFFOLD}~\citep{pmlr-v119-karimireddy20a} employs control variates to correct client drift, \textsc{FedNova}~\citep{fednova2020} normalizes local gradients to ensure consistent global updates, and \textsc{MOON}~\citep{li2021moon} aligns model representations across clients through contrastive learning.
Concept drift detection on edge devices, in turn, relies on lightweight, largely unsupervised techniques, including adaptive sliding windows~\citep{adwin2007,lulearningconceptdrift2019,aliadaptivedrift2018}, label-free monitoring of classification boundaries~\citep{d3_2019}, incremental statistical tests~\citep{reissmirnov2016,RAAB2020340}, and reservoir computing models~\citep{Gama2014,yamada2023}.\looseness=-1

\paragraph{FL for data streams\nodot} addresses communication efficiency, asynchronous collaboration, and unlabeled data handling, challenges orthogonal to drift-aware sample acquisition.
\textsc{ASO-Fed}~\citep{chenasofed2020} enables wait-free updates from streaming clients, \textsc{Fed-HIST}~\citep{zhang2024} leverages historical knowledge during streaming updates, and~\tcite{marfoq2023} formalizes FL for data streams via weighted empirical risk minimization.
Closer to our setting, \textsc{LeaDQ}~\citep{sun2025} queries informative samples from unlabeled streams via multi-agent reinforcement learning; in contrast to our sampler, it requires training decentralized query policies rather than scoring samples directly by feature geometry.
Finally, \ifanon\else our previous uncertainty-informed sampling approach~\citep{roedersampling2024} selects diverse batches under labeling constraints; the present work builds on it,\fi{} integrating diversity-driven acquisition into the federated training loop and coupling it with server-guided model updates under drift.\looseness=-1

\paragraph{Federated continual learning\nodot} (\textbf{FCL}) tackles catastrophic forgetting when client data arrives as a sequence of tasks.
\textsc{FedWeIT}~\citep{yoon2021federated} decomposes model weights into global and sparse task-specific parameters to limit inter-client interference, \textsc{CFeD}~\citep{ma2022cfed} distills knowledge of previous tasks through surrogate data, \textsc{FedKNOW}~\citep{luopan2023fedknow} integrates signature task knowledge on edge devices, and \textsc{FedSSI}~\citep{li2024fedssi} regularizes local updates via synaptic intelligence without rehearsal buffers.
Closest to our setting, uncertainty-aware memory management~\citep{serra2024onlinefcl} extends FCL to online streams without task boundaries.
These methods are \emph{model-centric}: they mitigate forgetting after the training data has been collected, assuming labeled task sequences and unconstrained data acquisition.
\textsc{STORMGUIDE-FL} instead operates \emph{upstream} of local training, deciding which samples from an unlabeled, unbounded stream are acquired and labeled under a fixed budget; its mined training sets could serve as input to any FCL method, rendering the two paradigms complementary.\looseness=-1

\paragraph{\textbf{Research Gap}} Most FL approaches assume stationary distributions or treat concept drift and client drift in isolation, drift detection methods target centralized single-client settings~\citep{lulearningconceptdrift2019}, and work on FL for data streams and FCL focuses on model-side adaptation, leaving open which samples to acquire from the stream in the first place.
To the best of our knowledge, no method jointly addresses both forms of drift at the level of resource-constrained data acquisition.

\section{Methodology}
\label{sec:methodology}
This section comprehensively outlines key design considerations of our approach:
Section~\ref{subsec:methodology_setup} establishes the foundational requirements of \textsc{STORMGUIDE-FL} by introducing the theoretical formulation of FL in non-IID streaming environments, including key definitions and notations as well as the problem setup. 
In Section~\ref{subsec:methodology_distributed_mining}, we discuss the general flow of data between the FL server and the participating FL clients and how our method integrates with existing frameworks.
Subsequently, Section~\ref{subsec:methodology_stage_1} outlines the sampling strategy deployed on FL clients to efficiently mine relevant data points from the observed data stream for further processing.
Section~\ref{subsec:methodology_stage_2} completes the client-side pipeline by describing how the mined samples are consumed by remote-guided local model updates that embed the server's guidance variate into every training step, and Section~\ref{subsec:methodology_server_control} concludes one round of FL training by describing the update procedure of the global model parameters and guidance variates on the FL server.
We summarize the frequently used notations in \appendixname~\ref{appendix:notation}.
\subsection{Problem Statement}
\label{subsec:methodology_setup}
We consider an FL environment as outlined in \figurename~\ref{fig:main_overview}, where the server $\mathcal{S}$ maintains the global model parameters~$\theta_{\mathcal{S}}$,
communicating with a set of clients $\mathcal{C}_1 \dots \mathcal{C}_N$ that maintain their local model parameters $\theta_{i}$, respectively.
Without loss of generality, the underlying deep neural network is simplified to a composition of feature extractor $f$ and classifier $g$, such that $f(\cdot ; \theta), g(\cdot ; \theta)$.
Furthermore, each client $\mathcal{C}_i$ has exclusive access to its unlabeled data stream $\mathcal{U}_i = \left\{ x^t_i \right\}_{t=1}^{T_i}$.
To reflect the motivational challenge of \emph{dual client heterogeneity}, as outlined in Section~\ref{subsec:intro_background} and illustrated in \figurename~\ref{fig:motivation}, we assume that the data observed across clients is inherently imbalanced and heterogeneous: the stream lengths $T_i$ may vary across clients, and both the cross-client and the within-client statistics of the observations are permitted to diverge.
We formalize the latter two conditions as follows.
\begin{definition}[Dual client heterogeneity]\label{def:dual_heterogeneity}
Let $\mathbb{P}_i^t(X)$ denote the marginal distribution governing the observation $x^t_i \in \mathcal{U}_i$ at timestep $t$, with each $x^t_i \sim \mathbb{P}_i^t$ drawn independently, let $\bar{\mathbb{P}}_i = \frac{1}{T_i}\sum_{t=1}^{T_i} \mathbb{P}_i^t$ denote the time-averaged stream distribution of client $\mathcal{C}_i$, and let $d(\cdot, \cdot)$ be a divergence on distributions (e.g., total variation).
The federation exhibits \emph{dual client heterogeneity} if
\begin{align*}
\text{(i)}\;\; & \Delta_{\mathrm{space}} \coloneqq \max_{i \neq j}\, d\big(\bar{\mathbb{P}}_i, \bar{\mathbb{P}}_j\big) > 0 && \text{\emph{(inter-client)}},\\
\text{(ii)}\;\; & \Delta_{\mathrm{time}} \coloneqq \max_{i}\, \max_{t \neq t^{\prime}}\, d\big(\mathbb{P}_i^t, \mathbb{P}_i^{t^{\prime}}\big) > 0 && \text{\emph{(intra-client)}},
\end{align*}
i.e., if spatial skew across clients and temporal drift within clients are present \emph{simultaneously}.
\end{definition}
The two conditions map directly onto the two drift phenomena of Section~\ref{subsec:intro_background}: $\Delta_{\mathrm{space}} > 0$ yields non-identical local objectives and hence \emph{client drift} at the update level~\citep{pmlr-v119-karimireddy20a}, while $\Delta_{\mathrm{time}} > 0$ is the defining condition of \emph{concept drift} that any one-pass acquisition strategy must track; the labeling budget $k_{\mathcal{S}}$ introduced below couples the two, since the samples acquired under (ii) determine the local updates aggregated under (i).
In our experiments, the Dirichlet concentration parameter $\alpha$ directly modulates $\Delta_{\mathrm{space}}$, while the streaming protocol sets the temporal structure of (ii); the empirical study of Section~\ref{sec:experiments} thus varies both conditions of Definition~\ref{def:dual_heterogeneity} independently.
Moreover, the drift regimes evaluated in this work arise as structural special cases of the sequence $(\mathbb{P}_i^t)_{t}$:
\emph{gradual} drift bounds each step, $d(\mathbb{P}_i^t, \mathbb{P}_i^{t+1}) \leq \epsilon$ for all $t$; \emph{sudden} drift concentrates the change in a single point, $\mathbb{P}_i^t = \mathbb{P}_{A}$ for $t < t^{\ast}$ and $\mathbb{P}_{B}$ thereafter; and \emph{recurring} drift imposes periodicity, $\mathbb{P}_i^{t+\tau} = \mathbb{P}_i^t$ with period $\tau$.
Gradual drift constitutes the default streaming protocol of our experiments.%

In order to account for realistic constraints on supervision, it is posited that a client may obtain ground-truth classification for at most $k_{\mathcal{S}} > 0$ data points drawn directly from its local data stream in each FL round, where the labeling budget $k_{\mathcal{S}}$ is announced by the server.
Additionally, label selection must be performed \emph{online}, meaning the client has to decide whether to request a label for each incoming data point immediately, without employing costly buffering strategies or revisiting previous stream observations.
Formally, in each round every participating client $\mathcal{C}_i$ selects a subset $D_i \subseteq \mathcal{U}_i$ with $|D_i| \leq k_{\mathcal{S}}$, where the inclusion decision for observation $x^t_i$ may depend only on the stream prefix $x^1_i, \dots, x^t_i$.
The primary global task in the context of FL optimization in this work is to efficiently find model parameters $\theta_\mathcal{S}$ that yield high prediction accuracy in centralized evaluation.

\subsection{\textsc{STORMGUIDE-FL}: Providing Global Learning Directions}
\label{subsec:methodology_distributed_mining}
\textsc{STORMGUIDE-FL} counters both concept drift and client drift by anchoring what each client samples and how each client trains to a shared reference point that the FL server updates and distributes every round.
Concretely, the server's current model parameters $\theta_{\mathcal{S}}$ ground the representation space in which clients identify diverse and informative samples from their local data streams (Section~\ref{subsec:methodology_stage_1}), while a guidance variable steers the local parameter updates computed on those samples toward the broader learning objective (Section~\ref{subsec:methodology_stage_2}), addressing the challenge of \emph{dual client heterogeneity} (Definition~\ref{def:dual_heterogeneity}; \figurename~\ref{fig:main_overview}, \emph{Data View}).
As part of the optimization strategy, we build upon Stochastic Controlled Averaging~\citep{pmlr-v119-karimireddy20a}, a variance-reduction method designed to correct client drift by maintaining auxiliary control variates.
Our approach analogously  tracks a server remote guidance variate $\mathbf{c}_{\mathcal{S}}$, representing the global update direction, and client remote guidance variates $\mathbf{c}_{i}$, which adjust local training dynamics to better conform with global consensus.
These variates are updated iteratively and serve to reduce inconsistency across client updates, particularly in non-IID settings.
Initially, the global model parameters $\theta_{\mathcal{S}}$ are either bootstrapped randomly or initialized with pretrained weights to leverage prior knowledge, and $\mathbf{c}_{\mathcal{S}}$ is safely set to 0~\citep{pmlr-v119-karimireddy20a}.\looseness=-1

During each communication round, the server transmits the global model parameter $\theta_{\mathcal{S}}$ and the server remote guidance variate $\mathbf{c}_{\mathcal{S}}$ to a randomly selected subset of clients $N^{\text{sub}} \subseteq \{1, \dots, N\}$.
Each participating client $i \in N^{\text{sub}}$ initializes its local model with the received server parameters, i.e. $\theta_i \gets \theta_{\mathcal{S}}$, and stores $\mathbf{c_{\mathcal{S}}}$ into temporary memory for use during the guided local model updates (see Section~\ref{subsec:methodology_stage_2}), rendering the client device ready to start the environment sensing.

\subsection{One-Pass Diversity Sampling for FL Data Streams}
\label{subsec:methodology_stage_1}
As soon as client $\mathcal{C}_i$ receives the training initiative from the server $\mathcal{S}$ for the current FL round, its first task is to employ a query strategy on the data stream $\mathcal{U}_i$ and select $k_{\mathcal{S}}$ data points for the next stage.
In general, these query strategies for streaming data have advanced to address the challenge of selecting the most informative samples from continuous data flows.
In the context of FL streaming data, the simplest approaches are \emph{stream-uniform sampling}~\citep{Haas2016streamsampling}, which accepts instances with uniform probability regardless of their content, and \emph{stream-interval sampling}, which selects every $n^{th}$ sample from the stream mechanically\footnote{There are more methods, but this paper covers the ones that are most commonly used on edge devices in real-world deployments.}~\citep{hus2012surveystreaming}.
Although these baseline methods are computationally straightforward and do not require on-device model evaluation, they are not able to leverage any insights about sample informativeness, often resulting in redundant selections and missed opportunities for capturing critical edge cases, especially in FL settings, where employing uninformed sampling strategies risk over-representing locally dominant patterns while failing to capture samples that are essential for robust global model performance.\looseness=-1

In this work, we adopt and extend volume-sampling techniques for streaming data~\citep{saran2023}, that intelligently select samples based on their contribution to model learning, while maintaining the computational efficiency and numerical stability required for real-time, large-scale streaming applications on FL devices~\citep{roedersampling2024}.
Hence, at this stage, our solution operates entirely in the embedding space of the underlying neural network (see \figurename~\ref{fig:main_overview}, \emph{Intuition View}), eliminating the need for labels during the sampling decision.
The method leverages the learned representation $f_i(x^t_i)$ of current stream observation $x^t_i$ to assess sample informativeness based solely on feature geometry, making instantaneous decisions possible without any labeling overhead.
The embedding-based nature of this strategy implies that the FL client can mine thousands of unlabeled streaming samples, whilst accepting only the $k_{\mathcal{S}}$ most informative instances as determined by their contribution to the feature space coverage.

\begin{algorithm}[t!] \small
    \caption{\small \textbf{O}ne-\textbf{P}ass \textbf{D}iversity \textbf{S}ampling for \textbf{FL} Data Streams \label{alg:fl_diversity_sampling}}
    \begin{algorithmic}[1]
        \Require Feature extractor $f_i(\cdot)$ on client $\mathcal{C}_i$, single observation $x^t_i$ sampled from  data stream $\mathcal{U}_i$ at timestep $t$, labeling budget $k_{\mathcal{S}}$, expected stream length $|\mathcal{U}_i|$
        \Ensure Acceptance decision \texttt{is\_accept}
        \State Initialize once at $t = 0$:
        \State \hspace*{\algorithmicindent} $(\hat{\mathbf{\Sigma}}_i^{0})^{-1} = \lambda^{-1} \mathbf{I}_{d \times d}$ \Comment{regularized by $\lambda$ for stability}
        \State \hspace*{\algorithmicindent} $\mathbf{A}^0_i = \mathbf{0}_{d\times d}$ \Comment{covariance over all data}
        \State \hspace*{\algorithmicindent} $n_i^0 = 0$ with $n_i^t \leq k_{\mathcal{S}}$ upon additions %
        \State \hspace*{\algorithmicindent} $\mathbf{L}^0_i$ $\gets$ \texttt{\textsc{cholesky}}($\hat{\mathbf{\Sigma}}_i^{0}$) \Comment{triangular factorization}
        \State $\mathbf{A}^t_i \leftarrow \frac{t-1}{t} \mathbf{A}^{t-1}_i + \frac{1}{t} f_i(x^t_i) f_i(x^t_i)^\top$ \Comment{covariance update}
        \State $q^t_i \gets (k_{\mathcal{S}} - n^t_i) / (|\mathcal{U}_i| - t)$ \Comment{see Eq.~\eqref{eq:adaptive_sampling_rate}}
        \State $p^t_i \gets q^t_i \cdot f_i(x^t_i)^\top (\hat{\mathbf{\Sigma}}_i^{t})^{-1} f_i(x^t_i) \ \text{tr}\left((\hat{\mathbf{\Sigma}}_i^{t})^{-1} \mathbf{A}^t_i\right)^{-1}$ \Comment{see Eq.~\eqref{eq:prob_update}}
        \State \textbf{with probability} $\min(p^t_i, 1.0)$:
        \State \hspace*{\algorithmicindent}$\mathbf{L}^{t+1}_i \gets \texttt{\textsc{chol\_lowrank}}\left(\mathbf{L}^t_i, f_i(x^t_i)\right)$ \Comment{see Eq.~\eqref{eq:chol_update}}
        \State \hspace*{\algorithmicindent}$(\hat{\mathbf{\Sigma}}_i^{t+1})^{-1} \leftarrow \left(\mathbf{L}^{t+1}_i (\mathbf{L}_i^{t+1})^\top\right)^{-1}$
        \State \hspace*{\algorithmicindent}$n_i^t \gets n_i^t + 1$
        \State \hspace*{\algorithmicindent}\texttt{is\_accept} $\leftarrow$ \textbf{True}
        \State \textbf{else}:
        \State \hspace*{\algorithmicindent}$(\hat{\mathbf{\Sigma}}_i^{t+1})^{-1}  \leftarrow (\hat{\mathbf{\Sigma}}_i^{t})^{-1} $
        \State \hspace*{\algorithmicindent}\texttt{is\_accept} $\gets$ \textbf{False}
    \State \Return \texttt{is\_accept}, $f_i(x^t_i)$
    \end{algorithmic}
\end{algorithm}

Technically, \emph{One-Pass Diversity Sampling for FL Clients} (\textbf{OPDS-FL}) is described as follows:
Given the data point $x^t_i$ sensed from the data stream $\mathcal{U}_i$ on the client $\mathcal{C}_i$ at time $t$, the algorithm extracts features $f_i(x^t_i) \in \mathbb{R}^d$, where $d$ denotes the size of the embedding dimension of the underlying model.
By continuously tracking a running covariance matrix over all data points $\mathbf{A}^t_i \in \mathbb{R}^{d\times d}$, the feature distribution over already observed stream samples is captured by\looseness=-1
\begin{equation}\label{eq:covariance_tracking}
    \mathbf{A}^t_i = \underbrace{\frac{t-1}{t} \mathbf{A}^{t-1}_i}_{\text{weighted previous covariance}} + \underbrace{\frac{1}{t} f_i\left(x^t_i\right) f_i\left(x^t_i\right)^\top}_{\text{new sample contribution}}.
\end{equation}
The sampling decision mainly relies on computing an acceptance probability that balances the remaining budget with the contribution of the sample $x^t_i$ to the coverage of the feature space. %
Therefore, the adaptive sampling rate $q^t_i$ is dynamically adjusted based on the remaining budget and expected stream length with
\begin{equation}\label{eq:adaptive_sampling_rate}
    q^t_i = \frac{k_{\mathcal{S}} - n^t_i}{|\mathcal{U}_i| - t},
\end{equation}
where $n_i^t$ is the count of selected samples up to timestep $t$ that is incremented whenever a sample is accepted.
The acceptance probability $p^t_i$ subsequently quantifies the determinantal contribution~\citep{saran2023} of the sample under inspection relative to the current feature covariance with
\begin{equation} \label{eq:prob_update}
    p^t_i = \frac{q^t_i \cdot f_i(x^t_i)^\top (\hat{\mathbf{\Sigma}}^{t}_i)^{-1} f_i(x^t_i)}{\text{tr}\left( \frac{1}{t} (\hat{\mathbf{\Sigma}}^{t}_i)^{-1} \sum_{j=1}^{t} f_i(x^j_i) f_i(x^j_i)^\top \right)},
\end{equation}
where $\hat{\mathbf{\Sigma}}^{t}_i$ denotes the covariance over all data points that have already been accepted for moving to the next mining phase.
A critical challenge at that point is to be computationally effective while maintaining numerical stability when updating the inverse of $\hat{\mathbf{\Sigma}}^{t}_i$ over potentially thousands of iterations.
Following the empirically validated solution proposed in~\ifanon\else our previous work~\citep{roedersampling2024}\fi, this problem is thoroughly addressed by maintaining a \emph{Cholesky factor} $\mathbf{L}^t_i$
over the course of the environmental sensing, and performing computationally cheap rank-one updates directly on the triangular factor: upon acceptance, $\mathbf{L}^{t+1}_i \gets \texttt{\textsc{chol\_lowrank}}\big(\mathbf{L}^t_i, f_i(x^t_i)\big)$ is the unique lower-triangular factor satisfying\looseness=-1
\begin{equation} \label{eq:chol_update}
    \mathbf{L}^{t + 1}_i \big(\mathbf{L}^{t + 1}_i\big)^\top = \mathbf{L}^t_i \big(\mathbf{L}^t_i\big)^\top + f_i(x^t_i)\, f_i(x^t_i)^\top,
\end{equation}
computed in $\mathcal{O}(d^2)$ operations without ever forming $\hat{\mathbf{\Sigma}}^{t}_i$ explicitly.
The key insight at this point is that instead of repeatedly inverting the covariance matrix $\hat{\mathbf{\Sigma}}_i^{t}$ at each timestep, maintaining the Cholesky factor allows us to compute the required inverse $(\hat{\mathbf{\Sigma}}_i^{t+1})^{-1}$ by calculating $\left(\mathbf{L}^{t+1}_i (\mathbf{L}_i^{t+1})^\top\right)^{-1}$ efficiently through sequential triangular system solves~\citep{seeger2004}.
This approach not only reduces computational complexity compared to direct matrix inverse calculations, it also provides superior numerical stability compared to the original low-rank update approach based on \emph{Sherman-Morrison-Woodbury formula}~\citep{saran2023, max1950inverting}, thus additionally lowering the propagation of round-off errors through the system that would otherwise render stream sampling on FL devices inoperable.

Overall, the integrated sampling algorithm makes binary accept/reject decisions for each inspected data point, accepting with probability $\min(p^t_i,1.0)$ to ensure data points are selected proportionally to their determinantal contribution while respecting local budget constraints.
This probabilistic selection, combined with the numerically stable Cholesky updates, enables efficient single-pass processing of streaming data on resource-constrained FL clients.
Details are summarized in Algorithm~\ref{alg:fl_diversity_sampling}.
Finally, accepted samples are assigned labels through pseudo-labeling mechanisms~\citep{kage2025reviewpseudolabelingcomputervision} or oracle consultation~\cite{settles2009active} (based on the deployment context), with the annotated sample-label pair $(x^t_i, y^t_i)$ subsequently being added to the local training set $D_i$ for the remote-guided model update phase.\looseness=-1

\subsection{Remote-Guided Local Model Updates}
\label{subsec:methodology_stage_2}
\textsc{STORMGUIDE-FL} addresses the remaining challenge of keeping local training consistent with the global optimization trajectory. %
Server guidance already shapes \emph{which} observations enter $D_i$: sampling operates directly in the representation space of the server-distributed model parameters $\theta_{\mathcal{S}}$ (Section~\ref{subsec:methodology_stage_1}).
The local update stage adds a second, complementary signal: the server remote guidance variate $\mathbf{c}_{\mathcal{S}}$ received at the beginning of the round (Section~\ref{subsec:methodology_distributed_mining}) corrects every mini-batch gradient toward the global update direction~\citep{pmlr-v119-karimireddy20a}, turning the mined observations into a guided local refinement $\theta^{\star}_i$ of the global model.
Coordinating both signals within a single communication round couples data acquisition and drift correction, two concerns that classical methods treat separately (see \figurename~\ref{fig:main_overview}, \emph{Intuition View}).\looseness=-1

Given the mined local training set $D_i$ with $|D_i| \leq k_{\mathcal{S}}$, the client computes $\theta^{\star}_i$ by performing $K$ mini-batch gradient steps, each applying the guidance-corrected update
\begin{equation}\label{eq:guided_local_update}
    \theta_i \leftarrow \theta_i - \eta_l \left( \nabla\mathcal{L}(\theta_i; b) - \mathbf{c}_i + \mathbf{c}_{\mathcal{S}} \right),
\end{equation}
where $\eta_l$ is the local learning rate, $\mathcal{L}(\theta_i; b) = \frac{1}{|b|}\sum_{(x, y) \in b} \ell(\theta_i; x, y)$ the average loss over mini-batch $b \subseteq D_i$, and $\mathbf{c}_i$ the client remote guidance variate.
The correction $\mathbf{c}_{\mathcal{S}} - \mathbf{c}_i$ removes the client-specific drift component from every step, so even a heavily skewed mini-batch is pushed along the global consensus trajectory~\citep{pmlr-v119-karimireddy20a, li2020, zhao2018}.

Deploying control variates in this label-scarce streaming setting departs from their classical use in variance-reduced optimization~\citep{johnson2013svrg} and FL~\citep{pmlr-v119-karimireddy20a}, which presumes a fixed dataset that faithfully represents the underlying objective: $\theta^{\star}_i$ is instead trained on a small, budgeted, once-observed subset of a transient stream.
The quality of its guidance correction therefore hinges on the quality of the sampled data, motivating the joint design of the sampling and update procedures.
We preliminarily formalize this coupling in \appendixname~\ref{app:theory}.%

Finally, the client folds $\theta^{\star}_i$ back into its remote guidance variate, reusing the already computed parameter difference %
\begin{equation}\label{eq:client_variate_update}
    \mathbf{c}^{+}_i = \mathbf{c}_i - \mathbf{c}_{\mathcal{S}} + \frac{1}{K \eta_l}\left(\theta_{\mathcal{S}} - \theta^{\star}_i\right),
\end{equation}
and communicates the resulting deltas $\left(\Delta\theta_i, \Delta \mathbf{c}_i\right) = \left(\theta^{\star}_i - \theta_{\mathcal{S}}, \; \mathbf{c}^{+}_i - \mathbf{c}_i\right)$ back to the FL server.
Algorithm~\ref{alg:fl_guided_update} in \appendixname~\ref{app:algorithms} details the full client-side procedure.

\begin{widetable}[t!]
\centering
\caption{Diversity scores of FL-grade sampling methods, evaluated on varying Dirichlet-partitioned data streams generated from four natural and one medical benchmark dataset in one pass, for client populations $N\in\{10,20\}$ and labeling budget $k_{\mathcal{S}} = 100$. Mean and standard deviation\tableonestdnote{} over three random seeds; best result per setting in bold with a highlighted cell.}
{\small\resizebox{\textwidth}{!}{\definecolor{plotorange}{HTML}{E8A33D}
\definecolor{plotnavy}{HTML}{3B4A6B}
\begin{tabular}{llcccccccccc}
\toprule
Sampling method &   \text{  $\alpha$  } & \multicolumn{2}{c}{CIFAR-10} & \multicolumn{2}{c}{FashionMNIST} & \multicolumn{2}{c}{OrganAMNIST} & \multicolumn{2}{c}{CIFAR-100} & \multicolumn{2}{c}{SVHN} \\
 & & $N=10$ & $N=20$ & $N=10$ & $N=20$ & $N=10$ & $N=20$ & $N=10$ & $N=20$ & $N=10$ & $N=20$ \\
\cmidrule(lr){1-12}
\multirow{3}{*}{stream-baseline} & 0.8 & $0.83 \pm 0.01$ & $0.80 \pm 0.01$ & $0.83 \pm 0.01$ & $0.79 \pm 0.01$ & $0.77 \pm 0.01$ & $0.76 \pm 0.00$ & $0.81 \pm 0.00$ & $0.81 \pm 0.01$ & $0.77 \pm 0.00$ & $0.76 \pm 0.00$ \\
 & 0.5 & $0.76 \pm 0.02$ & $0.71 \pm 0.01$ & $0.71 \pm 0.01$ & $0.70 \pm 0.00$ & $0.70 \pm 0.01$ & $0.72 \pm 0.01$ & $0.78 \pm 0.00$ & $0.79 \pm 0.00$ & $0.69 \pm 0.00$ & $0.69 \pm 0.01$ \\
 & 0.1 & $0.46 \pm 0.00$ & $0.54 \pm 0.01$ & $0.45 \pm 0.01$ & $0.49 \pm 0.01$ & $0.55 \pm 0.00$ & $0.56 \pm 0.01$ & $0.75 \pm 0.01$ & $0.71 \pm 0.00$ & $0.55 \pm 0.02$ & $0.45 \pm 0.02$ \\
\cmidrule(lr){2-12}
\multirow{3}{*}{stream-interval} & 0.8 & \cellcolor{plotorange!40}$\textbf{0.84} \pm 0.00$ & $0.80 \pm 0.00$ & $0.83 \pm 0.00$ & $0.79 \pm 0.00$ & $0.78 \pm 0.00$ & $0.77 \pm 0.00$ & $0.81 \pm 0.00$ & $0.80 \pm 0.00$ & \cellcolor{plotorange!40}$\textbf{0.81} \pm 0.00$ & \cellcolor{plotorange!40}$\textbf{0.78} \pm 0.00$ \\
 & 0.5 & \cellcolor{plotorange!40}$\textbf{0.80} \pm 0.00$ & $0.71 \pm 0.00$ & $0.71 \pm 0.00$ & $0.70 \pm 0.00$ & \cellcolor{plotorange!40}$\textbf{0.73} \pm 0.00$ & \cellcolor{plotorange!40}$\textbf{0.73} \pm 0.00$ & \cellcolor{plotorange!40}$\textbf{0.79} \pm 0.00$ & $0.79 \pm 0.00$ & \cellcolor{plotorange!40}$\textbf{0.71} \pm 0.00$ & $0.69 \pm 0.00$ \\
 & 0.1 & $0.48 \pm 0.00$ & $0.54 \pm 0.00$ & $0.45 \pm 0.00$ & $0.49 \pm 0.00$ & $0.57 \pm 0.03$ & $0.56 \pm 0.00$ & $0.74 \pm 0.00$ & $0.73 \pm 0.02$ & $0.53 \pm 0.00$ & \cellcolor{plotorange!40}$\textbf{0.46} \pm 0.00$ \\
\cmidrule(lr){2-12}
\multirow{3}{*}{coreset} & 0.8 & $0.76 \pm 0.00$ & $0.73 \pm 0.00$ & $0.22 \pm 0.00$ & $0.35 \pm 0.00$ & $0.27 \pm 0.00$ & $0.35 \pm 0.00$ & $0.76 \pm 0.00$ & $0.76 \pm 0.00$ & $0.76 \pm 0.00$ & $0.76 \pm 0.00$ \\
 & 0.5 & $0.69 \pm 0.00$ & $0.67 \pm 0.00$ & $0.31 \pm 0.00$ & $0.29 \pm 0.00$ & $0.30 \pm 0.00$ & $0.33 \pm 0.00$ & $0.72 \pm 0.00$ & $0.74 \pm 0.00$ & $0.68 \pm 0.00$ & $0.70 \pm 0.00$ \\
 & 0.1 & $0.42 \pm 0.00$ & $0.46 \pm 0.00$ & $0.30 \pm 0.00$ & $0.21 \pm 0.00$ & $0.39 \pm 0.00$ & $0.35 \pm 0.00$ & $0.69 \pm 0.00$ & $0.65 \pm 0.00$ & $0.52 \pm 0.00$ & $0.45 \pm 0.00$ \\
\cmidrule(lr){2-12}
\multirow{3}{*}{OPDS-FL (ours)} & 0.8 & $0.83 \pm 0.01$ & \cellcolor{plotorange!40}$\textbf{0.81} \pm 0.00$ & \cellcolor{plotorange!40}$\textbf{0.84} \pm 0.01$ & \cellcolor{plotorange!40}$\textbf{0.80} \pm 0.01$ & \cellcolor{plotorange!40}$\textbf{0.79} \pm 0.00$ & \cellcolor{plotorange!40}$\textbf{0.78} \pm 0.01$ & \cellcolor{plotorange!40}$\textbf{0.82} \pm 0.00$ & \cellcolor{plotorange!40}$\textbf{0.81} \pm 0.00$ & $0.79 \pm 0.02$ & $0.78 \pm 0.01$ \\
 & 0.5 & $0.78 \pm 0.01$ & \cellcolor{plotorange!40}$\textbf{0.72} \pm 0.01$ & \cellcolor{plotorange!40}$\textbf{0.74} \pm 0.01$ & \cellcolor{plotorange!40}$\textbf{0.71} \pm 0.00$ & $0.72 \pm 0.02$ & $0.72 \pm 0.01$ & $0.78 \pm 0.00$ & \cellcolor{plotorange!40}$\textbf{0.79} \pm 0.00$ & $0.70 \pm 0.01$ & \cellcolor{plotorange!40}$\textbf{0.70} \pm 0.00$ \\
 & 0.1 & \cellcolor{plotorange!40}$\textbf{0.49} \pm 0.01$ & \cellcolor{plotorange!40}$\textbf{0.55} \pm 0.01$ & \cellcolor{plotorange!40}$\textbf{0.46} \pm 0.01$ & \cellcolor{plotorange!40}$\textbf{0.50} \pm 0.00$ & \cellcolor{plotorange!40}$\textbf{0.58} \pm 0.01$ & \cellcolor{plotorange!40}$\textbf{0.57} \pm 0.01$ & \cellcolor{plotorange!40}$\textbf{0.75} \pm 0.01$ & \cellcolor{plotorange!40}$\textbf{0.73} \pm 0.00$ & \cellcolor{plotorange!40}$\textbf{0.55} \pm 0.01$ & $0.46 \pm 0.01$ \\
\bottomrule
\end{tabular}
}}
\label{tab:exp_diversity_in_fl}
\end{widetable}
\subsection{Centralized Remote Guidance Variate Updates}
\label{subsec:methodology_server_control}
The final execution step required to culminate one round of FL training controlled by \textsc{STORMGUIDE-FL} is the parameter and remote guidance variate update on the FL server $\mathcal{S}$:
In each communication round, each client $\mathcal{C}_i$ in the sampled subset $N^{\text{sub}}$ returns its model parameter and remote guidance variate deltas $(\Delta\theta_i, \Delta \mathbf{c}_i)$.
The server mimics \emph{Stochastic Controlled Averaging} by computing the average update and consolidating the global model parameters and the server remote guidance variate for the upcoming FL round:
\begin{align}
(\Delta \theta_{\mathcal{S}}, \Delta \mathbf{c}_\mathcal{S}) &= \frac{1}{|N^{\text{sub}}|} \sum_{i \in N^{\text{sub}}}\left(\Delta \theta_{i}, \Delta \mathbf{c}_i \right), \label{eq:server_aggregate}\\
\theta_{\mathcal{S}} &= \theta_{\mathcal{S}} + \eta_{\mathcal{S}} \Delta \theta_{\mathcal{S}}, \quad \mathbf{c}_\mathcal{S} = \mathbf{c}_\mathcal{S} + \frac{|N^{\text{sub}}|}{N} \Delta \mathbf{c}_\mathcal{S}, \label{eq:server_update}
\end{align}
where $\eta_{\mathcal{S}}$ denotes the global learning rate (\tcite{pmlr-v119-karimireddy20a} provides
extensive hyperparameter evaluation) and $N$ the total number of clients.
In terms of communication costs, our method maintains the same bandwidth overhead as \textsc{SCAFFOLD}; since only parameter and variate deltas are exchanged, the transmitted updates further admit a formal $\varepsilon$-differential-privacy extension, detailed in \appendixname~\ref{app:privacy}.

Overall, the proposed two-stage client pipeline distinguishes our approach from existing methods that address data heterogeneity and streaming drift separately.

\section{Experiments}
\label{sec:experiments}
We conduct experiments on real-world image classification datasets to confirm the effectiveness of our methodology in mitigating the \emph{dual client heterogeneity} problem (the primary motivation of our work, Section~\ref{subsec:intro_background}).
Our main findings are: \textcircled{1}
OPDS-FL delivers the highest sample diversity in nearly all highly non-IID settings and scales effectively with client population size, rendering it a strong and practical choice for FL use cases; and
\textcircled{2} \textsc{STORMGUIDE-FL} substantially improves the robustness of FL optimization in these challenging real-world scenarios.

All experiments were set up on a machine with an Intel Xeon Silver 4208 CPU (32 cores) and four NVIDIA Tesla V100 GPUs (32GB) running Ray clusters~\citep{moritz2018ray} for parallel client training.

\subsection{Efficiency of One-Pass Diversity Sampling}
\label{subsec:exp_1}
\paragraph{Setup}
We first verify the diversity of the samples OPDS-FL mines from non-IID data streams, using academic benchmark datasets cast into resource-constrained FL scenarios:
each client $\mathcal{C}_i$ must identify \mbox{$k_{\mathcal{S}} = 100$} representative data points from its stream $\mathcal{U}_i$ in one pass.
To simulate \emph{concept drift} in $\mathcal{U}_i$, we follow best practice~\citep{saran2023} and introduce artificial drift by sorting data points according to their first principal component.

We evaluate two client population settings, \mbox{$N = 10$} and \mbox{$N = 20$}.
To ensure fair comparison, the FL training is performed without client sub-sampling, i.e. \mbox{$N^{\text{sub}} = N$}, in all experiments.
Data heterogeneity across clients is controlled by Dirichlet partitioning~\citep{hsu2019measuringeffectsnonidenticaldata, li_fl2021} with concentration parameter $\alpha$: moderate (\mbox{$\alpha = 0.8$}), high (\mbox{$\alpha = 0.5$}), and extreme (\mbox{$\alpha = 0.1$}).
The experiment runs on five benchmark datasets spanning three visual domains and 10 to 100 classes: \mbox{CIFAR-10} and \mbox{CIFAR-100}~\cite{krizhevsky2009learning}, \mbox{FashionMNIST}~\cite{xiao2017fashionmnist}, \mbox{SVHN}~\cite{netzer2011svhn}, and \mbox{OrganAMNIST}, a subset of MedMNIST2D~\cite{medmnistv2} with 58,850 abdominal CT images in eleven categories. Extended dataset details are provided in \appendixname~\ref{appendix:datasets}.
The underlying neural network is chosen to meet the complexity of the data (ResNet-18 unless otherwise stated) and the model parameters $\theta_{\mathcal{S}}$ are initialized with pre-trained weights.

OPDS-FL is compared with three other lightweight methods suited to IoT deployment: stream-uniform sampling, which selects $k_{\mathcal{S}}$ data points uniformly at random from the stream; stream-interval sampling, which samples data points at regular temporal intervals; and coreset selection~\citep{sener_coresetcnn_2018} for streaming, which aims to construct a compact representative subset of the original data distribution.
The diversity score is based on normalized entropy~\cite{cover2006elements} of the class distribution, calculated by
\begin{equation}\label{eq:entropy}
(-\sum_{c} p_c \log_2 p_c) / (\log_2(\min(c^{\text{tot}}, k_{\mathcal{S}}))),
\end{equation}
where $p_c$ is the relative frequency of class $c$ in the selected samples, $c^{\text{tot}}$ is the total number of possible classes in the stream, and $k_{\mathcal{S}}$ is the number of samples selected; the score ranges from 0.0 (no diversity) to 1.0 (maximum diversity with balanced class selection).

\paragraph{Results}
In the most important scenarios for non-IID FL, the high heterogeneity settings (Dirichlet $\alpha$ = 0.1), \textbf{our approach achieves the best diversity scores in nine out of ten dataset and client population settings}, and stays within one standard deviation of the best method in the remaining one (SVHN, $N=20$), as shown in Table~\ref{tab:exp_diversity_in_fl}; across all 30 evaluated settings, OPDS-FL ranks first in 21.
As data homogeneity increases (with higher $\alpha$ values), the gap in diversity scores between our sampling module and the other approaches decreases, indicating that while all methods benefit from more uniform data distributions, OPDS-FL still remains competitive.
We further hypothesize that OPDS-FL improves over the course of FL training, as more diverse samples and successive model fine-tuning reinforce each other.
The results also reveal a dependency on the alignment between pre-training and target domain: margins are larger on natural images (CIFAR-10, CIFAR-100, FashionMNIST) than on the medical domain (OrganAMNIST) and the noisy digit imagery of SVHN.
We further provide extended evaluations of OPDS-FL in terms of numerical stability and runtime analysis in low-resource environments as well as an exemplary
visualization of the diversity score calculation defined in Equation~\eqref{eq:entropy} in \appendixname ~\ref{appendix:sec_opds_ext}.\looseness=-1

\subsection{Efficient Outlier Sampling in FL Streams}
\label{subsec:leverage_synthetic}
\paragraph{Setup}
The diversity results above depend on how cleanly a real embedding space separates rare from dominant classes, a property no single dataset lets us control.
Here we isolate the sampling problem itself: a one-pass, budget-limited stream contains a handful of genuinely rare classes, and the sampler must capture them within a single labeling budget $k_{\mathcal{S}}$.
Each client observes a synthetic stream with two dominant classes of $120$ points each ($\mathcal{N}$, $\sigma{=}1.0$, $16$-dimensional) and eight well-separated minority ``outlier'' classes, each drawn independently.
Every method runs for one round with budget $k_{\mathcal{S}}{=}30$ on a federation of $N{=}150$ independently constructed clients; we report the per-federation mean over $20$ trials.
OPDS-FL is compared against \emph{stream-uniform} and \emph{stream-interval}, the two baselines that inspect each sample only once and therefore remain deployable in a genuine streaming setting.
Task difficulty is controlled by the outlier classes' collective share of the stream: $43\%$ (\emph{moderate}), $16\%$ (\emph{high}), and $5\%$ (\emph{extreme}).

\begin{spanfigure}[hbt!]
\centering
\includegraphics[width=\figthreewidth]{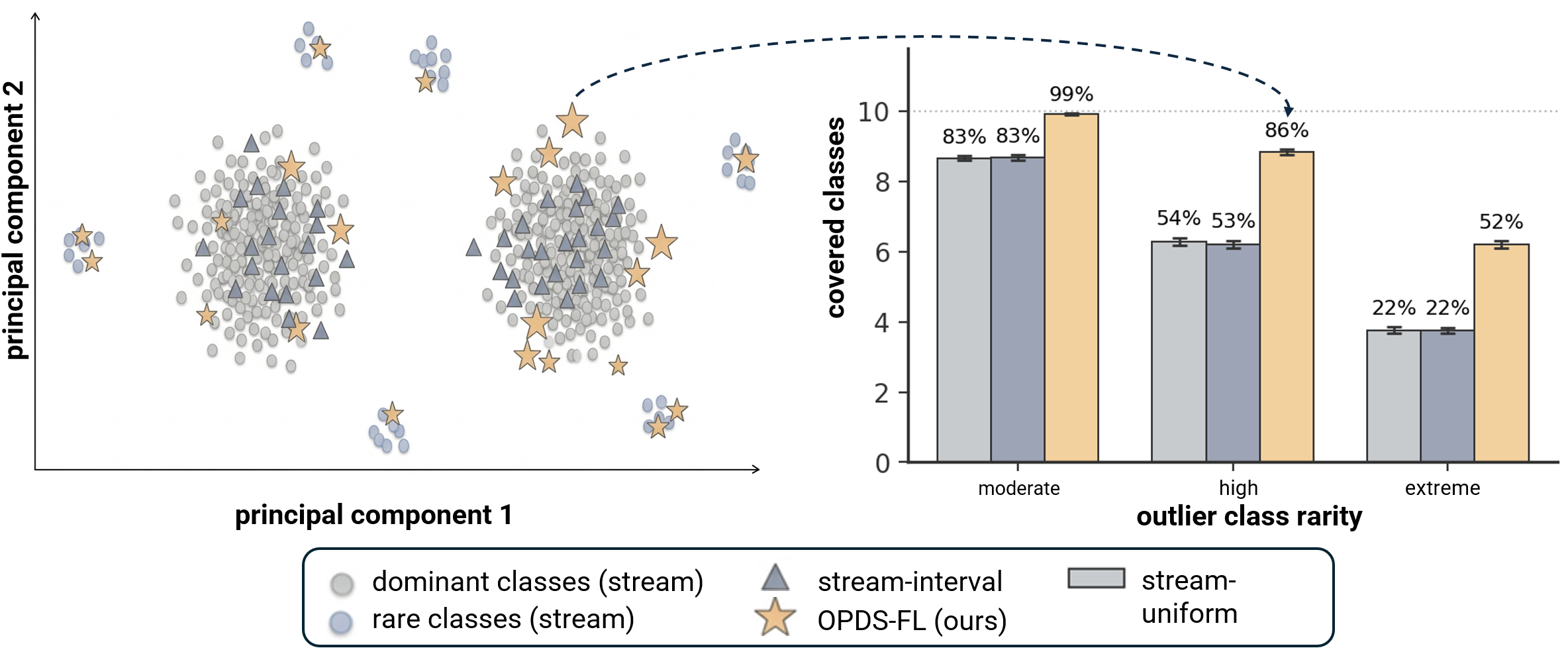}
\caption{Left: PCA view of one client's stream at the \emph{high} skew tier (gray: dominant classes, blue: rare classes), overlaid with the points selected by \textsc{Stream-interval} (triangles) and \textsc{OPDS-FL} (stars, sized by ridge-leverage score): only \textsc{OPDS-FL} reaches the isolated rare-class clusters. Right: mean classes covered (of $10$) per skew tier, with outlier hit-rates as \% labels; error bars show one s.d.\ over $20$ trials.}
\label{fig:leverage_skew_sweep}
\Description{A two-panel figure. The left panel is a 2D scatter plot with principal component 1 on the x-axis and principal component 2 on the y-axis, showing one client's synthetic data stream: two large dense clusters of gray dots (dominant classes) and eight small isolated clusters of blue dots (rare classes) scattered around the periphery. Overlaid are slate-blue triangles (Stream-interval picks), concentrated almost entirely inside the two dense gray clusters, and orange stars of varying size (OPDS-FL picks, sized by leverage score), which are scattered inside the dense clusters but also land directly on nearly every isolated rare-class cluster. The right panel is a grouped bar chart with three skew tiers on the x-axis (moderate, high, extreme, corresponding to 43, 16, and 5 percent of the stream being rare outlier classes) and covered classes on the y-axis, with a legend in a framed box below the chart. For each tier there are three bars: Stream-uniform (light gray) and Stream-interval (muted blue-gray) drop from about 8.7 classes covered (83 percent outlier hit-rate) at moderate skew, to 6.2-6.3 classes (53-54 percent) at high skew, to 3.7-3.8 classes (22 percent) at extreme skew. OPDS-FL (pastel orange) also declines but stays well above both baselines at every tier: 9.9 classes (99 percent) at moderate, 8.9 classes (86 percent) at high, and 6.2 classes (52 percent) at extreme skew.}
\end{spanfigure}

\paragraph{Results}
\figurename~\ref{fig:leverage_skew_sweep} shows that all methods degrade as the outlier classes become rarer, but at very different rates.
The one-pass baselines allocate their budget by a content-blind rule (pace-matched random acceptance or a fixed cadence), so their coverage is essentially a direct readout of the stream's class frequencies: the hit-rate falls from $83\%$ at moderate skew to $53$--$54\%$ at high and $22\%$ at extreme skew.
OPDS-FL instead spends its budget adaptively, favoring samples that are least represented among those already collected, and degrades far more gracefully ($99\%$, $86\%$, and $52\%$): at extreme skew it still matches the baselines' \emph{high}-skew hit-rate, although its outliers are about $3\times$ rarer ($5\%$ vs.\ $16\%$ of the stream).
Consequently, \textbf{the relative advantage of OPDS-FL grows monotonically with task difficulty}: a $1.2\times$ hit-rate multiplier at moderate skew widens to $1.6\times$ at high and $2.4\times$ at extreme skew.
The same qualitative pattern can be observed with real embeddings in Section~\ref{subsec:exp_1}: the advantage of diversity-seeking sampling grows precisely in the regime that matters most for FL, extreme client heterogeneity.
Here, the engineered stream makes the dominant/outlier structure unambiguous, confirming that this advantage is not an artifact of a particular dataset's embedding geometry.

\subsection{End-to-End FL Convergence Analysis}
\label{subsec:exp_2}

\begin{spantable}[t]
\centering
\caption{Convergence speed \emph{and} final accuracy on CIFAR-10 (from-scratch FedAvgCNN, $100$ rounds, identical streaming protocol). For each heterogeneity level we report the communication rounds to first reach the target accuracy $acc_t$ (lower is better; speedup over \textsc{FedAvg} in parentheses, red tick = \textsc{FedAvg} reference) and the final test accuracy (higher is better). Best per column in \textbf{bold}, second best \underline{underlined}; methods marked $^{*}$ use stream-interval sampling.}
\resizebox{\textwidth}{!}{\begin{tabular}{l r@{\hspace{0.3em}}l c r@{\hspace{0.3em}}l c r@{\hspace{0.3em}}l c}
\toprule
 Method & \multicolumn{3}{c}{$\alpha = 0.1\,(acc_t = 25\%)$} & \multicolumn{3}{c}{$\alpha = 0.5\,(acc_t = 30\%)$} & \multicolumn{3}{c}{$\alpha = 0.8\,(acc_t = 30\%)$} \\
\cmidrule(lr){2-4}\cmidrule(lr){5-7}\cmidrule(lr){8-10}
 & \multicolumn{2}{c}{Rounds} & Acc.\ (\%) & \multicolumn{2}{c}{Rounds} & Acc.\ (\%) & \multicolumn{2}{c}{Rounds} & Acc.\ (\%) \\
\midrule
\textsc{FedAvg}~\cite{mcmahan2017}$^{*}$ & 38 & \begin{tikzpicture}[baseline=-0.3ex, xscale=0.154]\fill[gray!20] (0,-0.02) rectangle (5.5,0.14);\fill[black] (0,0.02) rectangle (3.000,0.10);\fill[red] (3.0,-0.05) rectangle (3.08,0.17);\end{tikzpicture} \textcolor{darkgray}{(1.0\texttimes)} & 30.0 & 51 & \begin{tikzpicture}[baseline=-0.3ex, xscale=0.154]\fill[gray!20] (0,-0.02) rectangle (5.5,0.14);\fill[black] (0,0.02) rectangle (3.000,0.10);\fill[red] (3.0,-0.05) rectangle (3.08,0.17);\end{tikzpicture} \textcolor{darkgray}{(1.0\texttimes)} & 31.8 & 62 & \begin{tikzpicture}[baseline=-0.3ex, xscale=0.154]\fill[gray!20] (0,-0.02) rectangle (5.5,0.14);\fill[black] (0,0.02) rectangle (3.000,0.10);\fill[red] (3.0,-0.05) rectangle (3.08,0.17);\end{tikzpicture} \textcolor{darkgray}{(1.0\texttimes)} & 31.7 \\
\textsc{FedProx}~\cite{li2020}$^{*}$ & 38 & \begin{tikzpicture}[baseline=-0.3ex, xscale=0.154]\fill[gray!20] (0,-0.02) rectangle (5.5,0.14);\fill[black] (0,0.02) rectangle (3.000,0.10);\fill[red] (3.0,-0.05) rectangle (3.08,0.17);\end{tikzpicture} \textcolor{darkgray}{(1.0\texttimes)} & 30.0 & 51 & \begin{tikzpicture}[baseline=-0.3ex, xscale=0.154]\fill[gray!20] (0,-0.02) rectangle (5.5,0.14);\fill[black] (0,0.02) rectangle (3.000,0.10);\fill[red] (3.0,-0.05) rectangle (3.08,0.17);\end{tikzpicture} \textcolor{darkgray}{(1.0\texttimes)} & 31.8 & 62 & \begin{tikzpicture}[baseline=-0.3ex, xscale=0.154]\fill[gray!20] (0,-0.02) rectangle (5.5,0.14);\fill[black] (0,0.02) rectangle (3.000,0.10);\fill[red] (3.0,-0.05) rectangle (3.08,0.17);\end{tikzpicture} \textcolor{darkgray}{(1.0\texttimes)} & 31.8 \\
\textsc{MOON}~\cite{li2021moon}$^{*}$ & 38 & \begin{tikzpicture}[baseline=-0.3ex, xscale=0.154]\fill[gray!20] (0,-0.02) rectangle (5.5,0.14);\fill[black] (0,0.02) rectangle (3.000,0.10);\fill[red] (3.0,-0.05) rectangle (3.08,0.17);\end{tikzpicture} \textcolor{darkgray}{(1.0\texttimes)} & 29.8 & 51 & \begin{tikzpicture}[baseline=-0.3ex, xscale=0.154]\fill[gray!20] (0,-0.02) rectangle (5.5,0.14);\fill[black] (0,0.02) rectangle (3.000,0.10);\fill[red] (3.0,-0.05) rectangle (3.08,0.17);\end{tikzpicture} \textcolor{darkgray}{(1.0\texttimes)} & 31.9 & 62 & \begin{tikzpicture}[baseline=-0.3ex, xscale=0.154]\fill[gray!20] (0,-0.02) rectangle (5.5,0.14);\fill[black] (0,0.02) rectangle (3.000,0.10);\fill[red] (3.0,-0.05) rectangle (3.08,0.17);\end{tikzpicture} \textcolor{darkgray}{(1.0\texttimes)} & 31.8 \\
\textsc{SCAFFOLD}~\cite{pmlr-v119-karimireddy20a}$^{*}$ & \textbf{14} & \begin{tikzpicture}[baseline=-0.3ex, xscale=0.154]\fill[gray!20] (0,-0.02) rectangle (5.5,0.14);\fill[black] (0,0.02) rectangle (1.105,0.10);\fill[red] (3.0,-0.05) rectangle (3.08,0.17);\end{tikzpicture} \textcolor{darkgray}{(2.7\texttimes)} & \underline{36.1} & \underline{17} & \begin{tikzpicture}[baseline=-0.3ex, xscale=0.154]\fill[gray!20] (0,-0.02) rectangle (5.5,0.14);\fill[black] (0,0.02) rectangle (1.000,0.10);\fill[red] (3.0,-0.05) rectangle (3.08,0.17);\end{tikzpicture} \textcolor{darkgray}{(3.0\texttimes)} & 31.6 & \underline{18} & \begin{tikzpicture}[baseline=-0.3ex, xscale=0.154]\fill[gray!20] (0,-0.02) rectangle (5.5,0.14);\fill[black] (0,0.02) rectangle (0.871,0.10);\fill[red] (3.0,-0.05) rectangle (3.08,0.17);\end{tikzpicture} \textcolor{darkgray}{(3.4\texttimes)} & 30.1 \\
\midrule
\textsc{FedSSI}~\cite{li2024fedssi}$^{*}$ & 66 & \begin{tikzpicture}[baseline=-0.3ex, xscale=0.154]\fill[gray!20] (0,-0.02) rectangle (5.5,0.14);\fill[black] (0,0.02) rectangle (5.211,0.10);\fill[red] (3.0,-0.05) rectangle (3.08,0.17);\end{tikzpicture} \textcolor{darkgray}{(0.6\texttimes)} & 26.9 & 54 & \begin{tikzpicture}[baseline=-0.3ex, xscale=0.154]\fill[gray!20] (0,-0.02) rectangle (5.5,0.14);\fill[black] (0,0.02) rectangle (3.176,0.10);\fill[red] (3.0,-0.05) rectangle (3.08,0.17);\end{tikzpicture} \textcolor{darkgray}{(0.9\texttimes)} & 31.2 & 52 & \begin{tikzpicture}[baseline=-0.3ex, xscale=0.154]\fill[gray!20] (0,-0.02) rectangle (5.5,0.14);\fill[black] (0,0.02) rectangle (2.516,0.10);\fill[red] (3.0,-0.05) rectangle (3.08,0.17);\end{tikzpicture} \textcolor{darkgray}{(1.2\texttimes)} & 32.1 \\
\textsc{CFeD}~\cite{ma2022cfed}$^{*}$ & 63 & \begin{tikzpicture}[baseline=-0.3ex, xscale=0.154]\fill[gray!20] (0,-0.02) rectangle (5.5,0.14);\fill[black] (0,0.02) rectangle (4.974,0.10);\fill[red] (3.0,-0.05) rectangle (3.08,0.17);\end{tikzpicture} \textcolor{darkgray}{(0.6\texttimes)} & 25.8 & 62 & \begin{tikzpicture}[baseline=-0.3ex, xscale=0.154]\fill[gray!20] (0,-0.02) rectangle (5.5,0.14);\fill[black] (0,0.02) rectangle (3.647,0.10);\fill[red] (3.0,-0.05) rectangle (3.08,0.17);\end{tikzpicture} \textcolor{darkgray}{(0.8\texttimes)} & 32.1 & 63 & \begin{tikzpicture}[baseline=-0.3ex, xscale=0.154]\fill[gray!20] (0,-0.02) rectangle (5.5,0.14);\fill[black] (0,0.02) rectangle (3.048,0.10);\fill[red] (3.0,-0.05) rectangle (3.08,0.17);\end{tikzpicture} \textcolor{darkgray}{(1.0\texttimes)} & 32.2 \\
\textsc{FedAvg} + OPDS & \underline{34} & \begin{tikzpicture}[baseline=-0.3ex, xscale=0.154]\fill[gray!20] (0,-0.02) rectangle (5.5,0.14);\fill[black] (0,0.02) rectangle (2.684,0.10);\fill[red] (3.0,-0.05) rectangle (3.08,0.17);\end{tikzpicture} \textcolor{darkgray}{(1.1\texttimes)} & 33.0 & 42 & \begin{tikzpicture}[baseline=-0.3ex, xscale=0.154]\fill[gray!20] (0,-0.02) rectangle (5.5,0.14);\fill[black] (0,0.02) rectangle (2.471,0.10);\fill[red] (3.0,-0.05) rectangle (3.08,0.17);\end{tikzpicture} \textcolor{darkgray}{(1.2\texttimes)} & \underline{38.4} & 36 & \begin{tikzpicture}[baseline=-0.3ex, xscale=0.154]\fill[gray!20] (0,-0.02) rectangle (5.5,0.14);\fill[black] (0,0.02) rectangle (1.742,0.10);\fill[red] (3.0,-0.05) rectangle (3.08,0.17);\end{tikzpicture} \textcolor{darkgray}{(1.7\texttimes)} & \underline{39.2} \\
\midrule
\textsc{STORMGUIDE-FL} & \textbf{14} & \begin{tikzpicture}[baseline=-0.3ex, xscale=0.154]\fill[gray!20] (0,-0.02) rectangle (5.5,0.14);\fill[black] (0,0.02) rectangle (1.105,0.10);\fill[red] (3.0,-0.05) rectangle (3.08,0.17);\end{tikzpicture} \textcolor{darkgray}{(2.7\texttimes)} & \textbf{47.6} & \textbf{14} & \begin{tikzpicture}[baseline=-0.3ex, xscale=0.154]\fill[gray!20] (0,-0.02) rectangle (5.5,0.14);\fill[black] (0,0.02) rectangle (0.824,0.10);\fill[red] (3.0,-0.05) rectangle (3.08,0.17);\end{tikzpicture} \textcolor{darkgray}{(3.6\texttimes)} & \textbf{46.7} & \textbf{14} & \begin{tikzpicture}[baseline=-0.3ex, xscale=0.154]\fill[gray!20] (0,-0.02) rectangle (5.5,0.14);\fill[black] (0,0.02) rectangle (0.677,0.10);\fill[red] (3.0,-0.05) rectangle (3.08,0.17);\end{tikzpicture} \textcolor{darkgray}{(4.4\texttimes)} & \textbf{45.2} \\
\bottomrule
\end{tabular}
}
\label{tab:exp_convergence_in_fl}
\end{spantable}

\paragraph{Setup}
To evaluate the robustness and convergence behavior of \textsc{STORMGUIDE-FL}, this experiment tracks the communication rounds required to reach a per-setting target centralized test accuracy and the final accuracy after $100$ rounds on the CIFAR-10 classification task.
To isolate the federated optimizer and sampler from any pre-training confound, all methods train a compact convolutional network (FedAvgCNN, $0.88$M parameters) from scratch.
Data heterogeneity and concept drift follow the protocol of the first experiment: Dirichlet partitioning with $\alpha \in \{0.1, 0.5, 0.8\}$ and streams sorted by their first principal component.
\textsc{STORMGUIDE-FL} is compared to \textsc{FedAvg}, three state-of-the-art non-IID methods (\textsc{FedProx}, \textsc{SCAFFOLD}, \textsc{MOON}), and two federated continual-learning baselines (\textsc{FedSSI}, \textsc{CFeD}); streaming FL baselines address challenges orthogonal to drift-aware sample acquisition (asynchronous aggregation, parameter decomposition) and are therefore excluded.
Non-streaming methods receive their data points via industry-standard stream-interval sampling.
To reflect resource scarcity, each client samples at most $k_{\mathcal{S}} = 30$ data points per round.
All experiments use $N = 20$ clients with $N^{\text{sub}} = 15$ participating per round, each performing $5$ epochs of local training before sending its update to the FL server.

\paragraph{Results}
Table~\ref{tab:exp_convergence_in_fl} shows that \textsc{STORMGUIDE-FL} \textbf{is simultaneously the fastest to converge and the most accurate across every heterogeneity level} (tying \textsc{SCAFFOLD} on rounds-to-target at $\alpha=0.1$, but leading outright on accuracy at every level).
It reaches the per-setting target in only $14$ rounds for all three values of $\alpha$, corresponding to $2.7\times$, $3.6\times$, and $4.4\times$ speedups over \textsc{FedAvg}.
Its final accuracy of $47.6\%$, $46.7\%$, and $45.2\%$ ($\alpha = 0.1, 0.5, 0.8$) makes it the only method to surpass $45\%$, exceeding the strongest baseline by $+11.5$, $+8.3$, and $+6.0$ points.

The \emph{ablations} in Table~\ref{tab:exp_convergence_in_fl} localize the source of this gain.
\textsc{FedProx} and \textsc{MOON} are indistinguishable from vanilla \textsc{FedAvg} (within $0.3$ points everywhere), so proximal and contrastive corrections do not help under this streaming, drift-heavy regime; the gain cannot stem merely from a stronger optimizer.
Adding OPDS sampling to \textsc{FedAvg} alone yields $+3.0$ to $+7.5$ points across heterogeneity levels, and \textsc{SCAFFOLD}'s drift correction alone accelerates early convergence.
Only their combination reaches the $45\%$-plus regime, indicating a super-additive interaction between drift correction and diversity-driven sampling rather than two independent effects.

Finally, the strongest interval baseline exposes a failure mode that our sampler avoids.
\textsc{SCAFFOLD}+interval reaches the (low) accuracy targets quickly but does not sustain them: its accuracy peaks mid-training and then degrades (\figurename~\ref{fig:drift}, top).
Under identical drift correction, OPDS sampling shows no such degradation, which we attribute to the diversity-balanced batches keeping the remote guidance variates well conditioned.

\subsection{Robustness to Drift Regimes}
\label{subsec:drift}
\paragraph{Setup}
Beyond the PCA-sorted \emph{artificial} drift used as the default streaming protocol throughout this paper, we stress-test the streaming sampler under two further non-stationary regimes: a single \emph{sudden} distributional step, in which the local stream switches abruptly between two disjoint halves of the label space, and a \emph{recurring} (cyclic) drift, in which it alternates repeatedly between two opposing class orderings.
To separate the contributions of the client-side sampler and the server-side aggregator, we compare \textsc{FedAvg}+interval (weak baseline), \textsc{SCAFFOLD}+interval (strong baseline), and \textsc{STORMGUIDE-FL} on the from-scratch FedAvgCNN backbone over $100$ rounds at $\alpha \in \{0.1, 0.5, 0.8\}$ across three seeds.
\figurename~\ref{fig:drift} reports all three regimes, with the artificial-drift condition of Table~\ref{tab:exp_convergence_in_fl} re-run across the same seeds for a directly comparable band.\looseness=-1

\begin{spanfigure}[!htb]
\centering
\includegraphics[width=\figfourwidth]{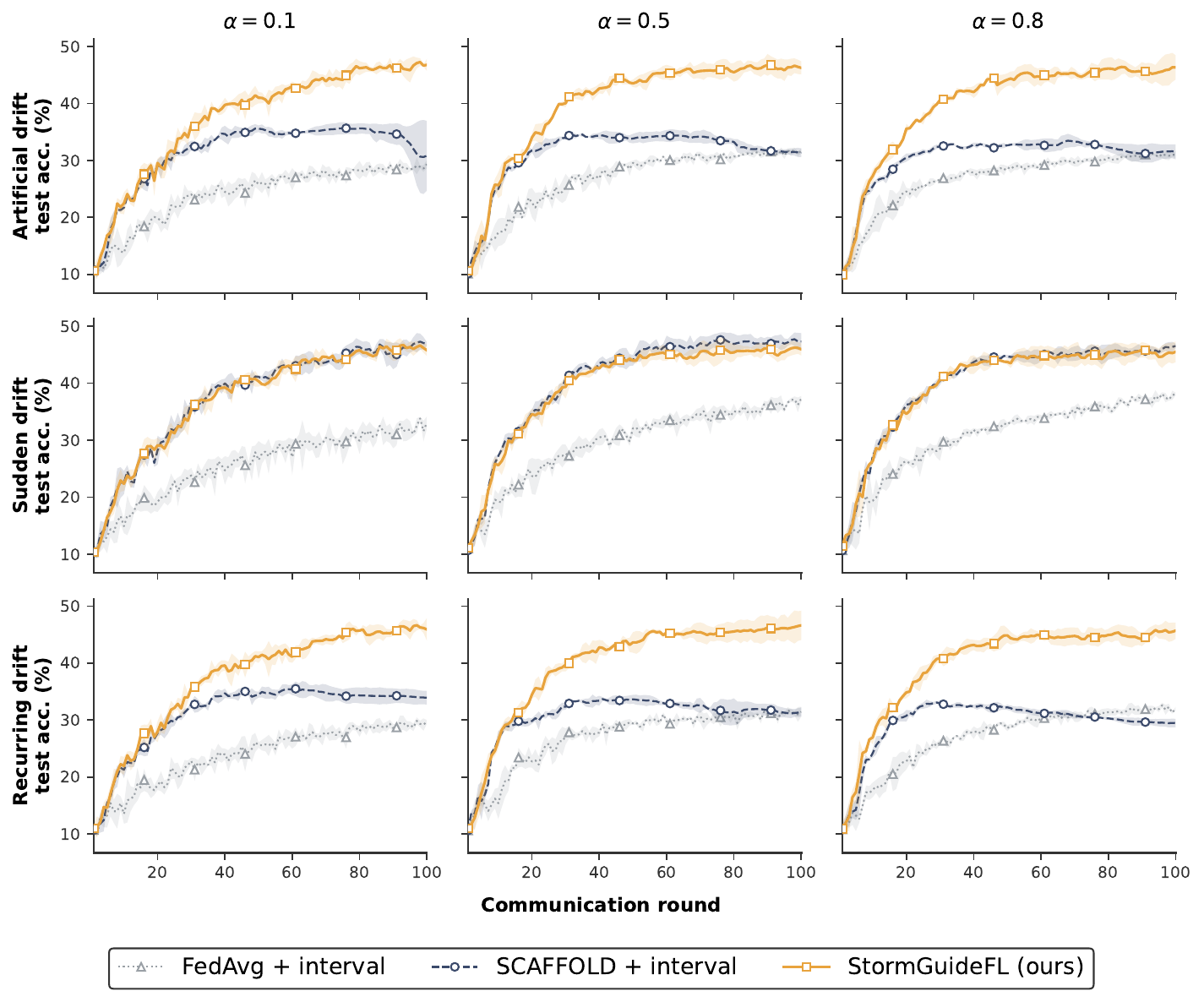}
\caption{Centralized test accuracy over $100$ rounds under artificial (top), sudden (middle), and recurring (bottom) drift, for $\alpha \in \{0.1, 0.5, 0.8\}$.}
\label{fig:drift}
\end{spanfigure}

\paragraph{Results}
Each regime stresses one component: \emph{sudden} drift the aggregator, \emph{recurring} drift the sampler.
Under sudden drift, \textsc{SCAFFOLD}'s control variates absorb the one-off shift ($46.5$--$47.3\%$), \textsc{FedAvg} trails by $8$--$14$ points, and the sampler choice is immaterial: OPDS stays within $1$--$1.4$ points of interval sampling.
Under recurring drift the ranking inverts: \textsc{STORMGUIDE-FL} retains $45.7$--$46.6\%$, whereas both interval baselines collapse to $29.5$--$33.9\%$ \emph{regardless of aggregator}, a $+12$ to $+16$ point margin.
Crucially, \textbf{only \textsc{STORMGUIDE-FL} is invariant to the drift regime} ($45.5$--$46.6\%$ throughout), while the strongest baseline swings by $13$--$17$ points between regimes.

The mechanism is the sampling objective: OPDS scores each unlabeled embedding by its leverage $\lVert u \rVert^{2}_{\Sigma^{-1}}$ against the covariance of the already-selected set, maximizing coverage of the \emph{feature-space} ellipsoid rather than label balance.
A fixed stride still spans a stream that shifts once, so interval sampling survives sudden drift unharmed.
A cyclic stream, however, \emph{aliases} that stride: the sampler revisits the same phases, entire feature directions stay uncovered even when labels appear balanced, and the model oscillates with the cycle.
The OPDS objective is order-invariant, so it recovers a representative cross-section of either stream; a coverage gap opened at sampling time cannot be closed by any aggregator, which is why the \textsc{SCAFFOLD} and \textsc{FedAvg} interval variants coincide under recurring drift.

\section{Conclusion}
\label{sec:conclusion}
We presented \textsc{STORMGUIDE-FL}, a federated optimization and data-mining framework for evolving non-IID data streams on edge devices.
Where existing methods treat data acquisition and drift correction as separate concerns, our approach aligns them within each communication round: diversified stream sampling and server-guided local updates jointly keep each client's refinement of the global model focused on the global optimum as its local data stream observations evolve.
\emph{Future work} includes client-side change detection that resets the running statistics, and a local gradient-alignment extension that carries the remote guidance from acquisition to the selection of already-labeled samples.
Extending the evaluation to further modalities and larger client populations is a natural next step.
We hope this line of research enables long-lived FL deployments on resource-limited edge fleets, from trail cameras monitoring changing wildlife habitats to medical imaging devices.\looseness=-1

\ifanon\else
\begin{credits}
\subsubsection{\ackname}
This work was funded in part by the European Regional Development Fund (ERDF) under grant FKZ:2404-003-1.2 (EU-EFRE GREEN-INNO), and in part by the Bavarian HighTech Agenda, specifically via the Würzburg Center for Artificial Intelligence (CAIRO) and the ProPere THWS scholarship.
MR would like to thank Valerie Vaquet for the fruitful discussion on the characterization of concept drift in FL.
\end{credits}
\fi

\bibliographystyle{icml2026}
\bibliography{main}

\clearpage
\appendix
\onecolumn
\setcounter{figure}{0}\renewcommand{\thefigure}{A\arabic{figure}}
\setcounter{table}{0}\renewcommand{\thetable}{A\arabic{table}}
\setcounter{algorithm}{0}\renewcommand{\thealgorithm}{A\arabic{algorithm}}
\setcounter{equation}{0}\renewcommand{\theequation}{A\arabic{equation}}
\section*{Appendix}
Due to space limitations in the main text, additional declarations, results and discussions are provided in the supplementary material, organized as follows:
\begin{itemize}
    \item We first summarize the notations frequently used in \appendixname~\ref{appendix:notation} to facilitate understanding of this work.
    \item We include additional information to support reproducibility, covering dataset descriptions, data splits, baseline model details, hyperparameter selection, and the full client-side update procedure (Algorithm~\ref{alg:fl_guided_update}) in \appendixname~\ref{appendix:reproduce}.
    \item We report extended OPDS-FL experiments in \appendixname~\ref{appendix:sec_opds_ext}, highlighting numerical stability and efficiency on resource-constrained federated clients. Furthermore, a comprehensive visualization of the diversity score calculation employed in the experiments is presented.
    \item We provide theoretical analysis in \appendixname~\ref{app:theory}, establishing a sampler-dependent convergence guarantee for the remote-guided local updates under bounded drift and quantifying how the diversity sampler controls that guarantee under a fixed labeling budget.
    \item We present a formal $\varepsilon$-differential-privacy extension of the transmitted updates in Appendix~\ref{app:privacy}, with noise calibrated to standard smoothness and bounded-gradient assumptions.
\end{itemize}

\section{Notation}
\label{appendix:notation}
\begin{table}[h]
    \caption{Notation Summary}
    \label{tab:notations}
    \centering\small
        \begin{tabular}{ll}
        \toprule
        \text {Notation   } & \text {Description} \\
        \hline
        \text{$N, N^{\text{sub}}$} & \text{total num. of clients, num. of sampled clients}\\
        \text{$\mathcal{S, C}_i$} & \text{server $\mathcal{S}$, client $\mathcal{C}_i, i = 0 \dots N$}\\
        \text{$\mathcal{U}_i$} & \text{non-IID data stream, observed on client $\mathcal{C}_i$}\\
        \text{$x$} & \text{stream observation, $x \in \mathcal{U}$}\\
        \text{$\theta$} & \text{model parameters}\\
        \text{$q$} & \text{adaptive sampling rate}\\
        \text{$p$} & \text{sampling probability}\\
        \text{$k_{\mathcal{S}}$} & \text{uniform labeling budget, announced by $\mathcal{S}$}\\
        \text{$D_i$} & \text{local training dataset mined by OPDS-FL}\\
        \text{$\theta^{\star}_i$} & \text{guided local refinement, trained on $D_i$}\\
        \text{$K, \eta_l, \eta_{\mathcal{S}}$} & \text{local steps, local/global learning rate}\\
        \text{$\mathbf{c}_{\mathcal{S}},\mathbf{c}_i$} & \text{server/client remote guidance variate}\\
        \bottomrule
        \end{tabular}
\end{table}

\section{Reproducibility}
\label{appendix:reproduce}
\subsection{Datasets}
\label{appendix:datasets}
\begin{itemize}
    \item \textbf{CIFAR-10}~\cite{krizhevsky2009learning}: This is a small dataset for general object recognition, containing ten categories of 32×32 RGB color images such as airplanes, automobiles, birds, cats, and ships.
    It includes 50,000 training samples and 10,000 test samples, making it a standard benchmark for evaluating image classification models and deep learning architectures in computer vision.
    \item \textbf{Fashion-MNIST}~\cite{xiao2017fashionmnist}: This is a benchmark dataset for image classification focused on fashion products.
    It contains ten categories of grayscale images (28×28 pixels) representing items such as shirts, shoes, and bags.
    The dataset consists of 60,000 training samples and 10,000 test samples, designed as a more challenging, drop-in replacement for the classic MNIST handwritten digits.
    Each image is centered and normalized to reduce pre-processing complexity, making it a standard testbed for evaluating basic and deep learning models on grayscale visual data.
    \item \textbf{OrganAMNIST}: This is a medical imaging dataset derived from the MedMNIST~\cite{medmnistv2} collection, based on 3D abdominal CT scans.
    It contains eleven categories corresponding to different abdominal organs (e.g., liver, spleen, kidneys, pancreas, etc.), converted into 2D grayscale slices (28×28 pixels) for classification tasks.
    We use the official MedMNIST~v2 release and merge its training, validation, and test splits into a single pool of 58,850 images, from which the federated partitions and client streams are constructed.
    OrganAMNIST is used to benchmark lightweight deep learning models for medical image recognition and to study transfer learning and domain adaptation in biomedical contexts.
    \item \textbf{CIFAR-100}~\cite{krizhevsky2009learning}: A fine-grained companion to CIFAR-10, containing 60,000 32×32 RGB color images evenly distributed over 100 object categories (50,000 training and 10,000 test samples).
    The tenfold increase in class count over CIFAR-10 renders class coverage under a fixed labeling budget substantially more challenging, making this dataset a suitable testbed for evaluating sampling diversity at scale.
    \item \textbf{SVHN}~\cite{netzer2011svhn}: The Street View House Numbers dataset consists of 32×32 RGB images of printed digits (ten classes) cropped from real-world house-number photographs.
    We use the standard training and test subsets, totaling 99,289 images.
    Its noisy capture conditions and frequent distractor digits at the image borders make SVHN a challenging benchmark for representation-based sampling strategies.
\end{itemize}

\subsection{Dataset Partitioning and Drift Simulation}
\label{appendix:distribution_drift}
\begin{figure}[!h]
    \centering
    \begin{subfigure}[b]{\figAonesub}
        \centering
        \includegraphics[width=\textwidth]{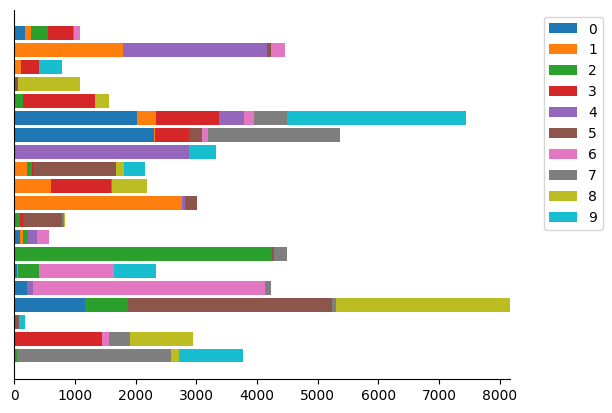}
        \caption{$\alpha = 0.1$}
        \label{fig:class_dist_01}
    \end{subfigure}
    \hfill
    \begin{subfigure}[b]{\figAonesub}
        \centering
        \includegraphics[width=\textwidth]{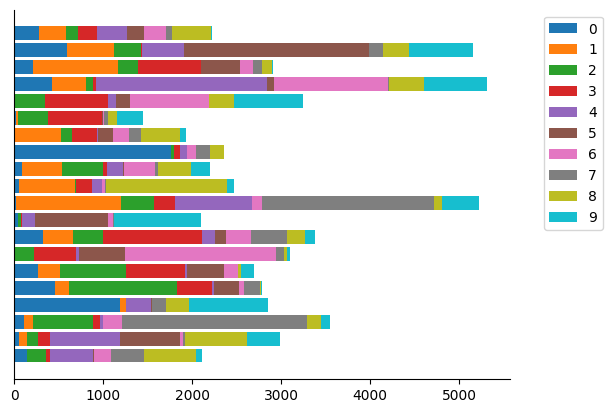}
        \caption{$\alpha = 0.5$}
        \label{fig:class_dist_05}
    \end{subfigure}
    \figAonebreak
    \begin{subfigure}[b]{\figAonesub}
        \centering
        \includegraphics[width=\textwidth]{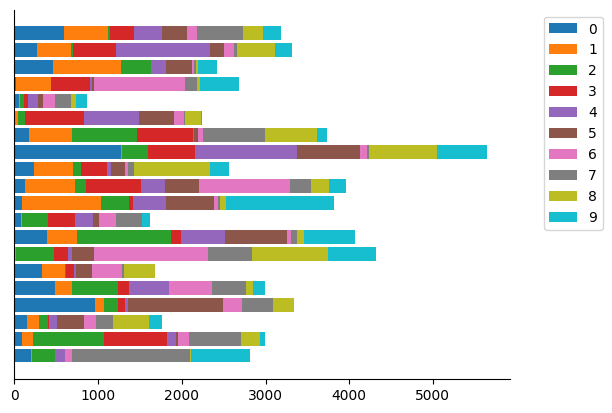}
        \caption{$\alpha = 0.8$}
        \label{fig:class_dist_08}
    \end{subfigure}
    
    \caption{Class distribution across 20 FL clients in CIFAR-10 training set under different Dirichlet distribution concentration parameters $\alpha$.
    Lower $\alpha$ values result in more heterogeneous (non-IID) data partitions, where each client obtains fewer classes with highly imbalanced distributions.
    Higher $\alpha$ values lead to more balanced class distributions across clients.
    The x-axis denotes the number of samples (with color-coded class membership) used to compose the data stream $\mathcal{U}_i$, the y-axis denotes the client $\mathcal{C}_i$, respectively.}
    \label{fig:class_distributions}
\end{figure}

We simulate \emph{dual client heterogeneity} through two mechanisms: 
Data is partitioned across clients using Dirichlet distribution-based sampling with concentration parameter $\alpha \in \{0.1, 0.5, 0.8\}$, where lower values create stronger non-IID conditions.
\figurename~\ref{fig:class_distributions} provides a visual intuition on how $\alpha$ impacts the class distribution across FL clients in our experimental setup.

Moreover, to artificially induce \emph{concept drift} within the local stream $\mathcal{U}_i$ of each client $\mathcal{C}_i$, data points are sorted by their first principal component, creating gradual distributional shifts.
This setup captures both spatial heterogeneity (different distributions across clients observing different scenes) and temporal heterogeneity (evolving distributions within each client caused by sensor degradation, seasonal effects, etc. ), reflecting the dual drift challenges in real-world federated deployments.
    
\subsection{Model Baselines}
\label{appendix:baselines}
\begin{itemize}
    \item \textsc{FedAvg}~\cite{mcmahan2017}: A distributed optimization algorithm where multiple clients perform local gradient descent on their private datasets for several epochs, after which a central server aggregates the updated model parameters through weighted averaging proportional to dataset sizes.
    This synchronized global model is then redistributed to all clients, and the process iterates until convergence.
    \item \textsc{FedProx}~\cite{li2020}: Extends \textsc{FedAvg} by adding a proximal term to the local objective function, limiting the divergence of local updates from the global model to better handle heterogeneous data distributions and system constraints across clients.
    \item \textsc{MOON}~\cite{li2021moon}: Model-Contrastive Federated Learning augments standard local training with a contrastive loss that aligns each client’s current model representations with those of the global model while pushing them away from outdated local models, thereby explicitly correcting client drift under non-IID data and improving robustness over vanilla \textsc{FedAvg}.
    \item \textsc{SCAFFOLD}~\cite{pmlr-v119-karimireddy20a} (Stochastic Controlled Averaging for Federated Learning): Introduces control variates that track the difference between local and global update directions, correcting for client drift caused by data heterogeneity and reducing the number of communication rounds required for convergence.
    \item \textsc{FedSSI}~\cite{li2024fedssi} (Synergistic Synaptic Intelligence): A rehearsal-free federated continual learning method that regularizes local updates by penalizing changes to parameters identified as important for previously seen objectives, consolidating this importance over time without requiring a replay buffer or labeled task identity.
    \item \textsc{CFeD}~\cite{ma2022cfed} (Continual Federated learning via Distillation): A federated continual learning method that mitigates catastrophic forgetting via knowledge distillation, normally using a shared surrogate dataset to distill knowledge of previous tasks into the updated local model.
\end{itemize}

\paragraph{Adapting FCL baselines to a boundary-free stream.}
\textsc{FedSSI} and \textsc{CFeD} assume labeled task sequences with known boundaries, and \textsc{CFeD} further assumes a persistent surrogate dataset for distillation; neither holds in our boundary-free, budget-constrained stream. We adapt both to run natively under this protocol: \textsc{FedSSI} treats every communication round as an implicit task boundary, consolidating its synaptic-intelligence importance continuously round-by-round against the global model received that round, preserving its rehearsal-free design; \textsc{CFeD} instead distills from each client's own model from the previous round (self-distillation on the current round's mined set $D_i$), substituting the missing surrogate dataset with a temporally-local teacher. Both use their native fixed-interval acquisition under the same budget $k_{\mathcal{S}}=30$ and \textsc{FedAvg}-style aggregation (\textsc{SCAFFOLD}'s control variates are not applied to their local steps). \textsc{FedWeIT}~\citep{yoon2021federated} and \textsc{FedKNOW}~\citep{luopan2023fedknow} are excluded, as their per-task weight masks and signature task knowledge presuppose task boundaries that would require an ad hoc, hard-to-interpret adaptation.

\paragraph{Adapting the coreset sampling baseline.}
The $k$-center greedy rule of coreset selection~\citep{sener_coresetcnn_2018} requires the full candidate pool, which a single pass over the stream never materializes.
We therefore run it as a streaming adaptation: a sample is accepted when its distance to the already selected set exceeds one half of the median pairwise distance among the ten most recent samples, with interval pacing before that estimate stabilizes.

\subsection{Hyperparameter Selection}
\label{appendix:hyperparameter}
We adopt hyperparameter settings informed by the extensive evaluations provided in \textsc{SCAFFOLD}, which systematically studied control variate-based federated optimization across diverse heterogeneity settings.
Following their findings, we set the global learning rate $\eta_\mathcal{S} = 1.0$ for server-side parameter aggregation (see Equation~\eqref{eq:server_update}) in our experiments.
OPDS-FL itself is hyperparameter-free: the labeling budget $k_{\mathcal{S}}$ is announced by the server, the expected stream length $|\mathcal{U}_i|$ is a property of the sensing period, and the acceptance probability adapts automatically through the tracked feature covariance (Section~\ref{subsec:methodology_stage_1}).
The initialization constant $\lambda$ serves numerical stability only.
No sampler parameter was tuned in any experiment reported in this paper.

\subsection{Client-Side Update Procedure}
\label{app:algorithms}
Algorithm~\ref{alg:fl_guided_update} details the client-side procedure summarized in Section~\ref{subsec:methodology_stage_2}: guidance-corrected local training on the mined set $D_i$, followed by the remote guidance variate update of Equation~\eqref{eq:client_variate_update} and the delta computation returned to the server.
\begin{algorithm}[htb] \small
    \caption{\small Remote-Guided Local Model Update for FL Clients\label{alg:fl_guided_update}}
    \begin{algorithmic}[1]
        \Require local training set $D_i$ mined on client $\mathcal{C}_i$ (see Algorithm~\ref{alg:fl_diversity_sampling}), global model parameters $\theta_{\mathcal{S}}$, server remote guidance variate $\mathbf{c}_{\mathcal{S}}$, client remote guidance variate $\mathbf{c}_i$, local steps $K$, local learning rate $\eta_l$
        \Ensure local differences $\Delta\theta_i$ and $\Delta \mathbf{c}_i$
        \State $\theta_i \gets \theta_{\mathcal{S}}$ \Comment{initialize from global model}
        \For{$k = 1, \dots, K$}
            \State sample mini-batch $b \subseteq D_i$
            \State $\mathbf{g}_i \gets \nabla\mathcal{L}(\theta_i; b)$ \Comment{mini-batch gradient}
            \State $\theta_i \gets \theta_i - \eta_l \left(\mathbf{g}_i - \mathbf{c}_i + \mathbf{c}_{\mathcal{S}}\right)$ \Comment{guided update, see Eq.~\eqref{eq:guided_local_update}}
        \EndFor
        \State $\theta^{\star}_i \gets \theta_i$ \Comment{final iterate is the guided local refinement}
        \State $\mathbf{c}^{+}_i \gets \mathbf{c}_i - \mathbf{c}_{\mathcal{S}} + \frac{1}{K\eta_l}\left(\theta_{\mathcal{S}} - \theta^{\star}_i\right)$ \Comment{see Eq.~\eqref{eq:client_variate_update}}
        \State $\left(\Delta\theta_i, \Delta\mathbf{c}_i\right) \gets \left(\theta^{\star}_i - \theta_{\mathcal{S}}, \mathbf{c}^{+}_i - \mathbf{c}_i\right)$ \Comment{deltas w.r.t.\ old $\mathbf{c}_i$}
        \State $\mathbf{c}_i \gets \mathbf{c}^{+}_i$ \Comment{persist for the next round}
    \State \Return $\Delta\theta_i, \Delta \mathbf{c}_i$
    \end{algorithmic}
\end{algorithm}

\section{Extended Experiments on OPDS-FL}
\label{appendix:sec_opds_ext}
\ifanon
\else In our previous work~\citep{roedersampling2024}, we thoroughly examined the benefits of the integrated sampling strategy for large-scale, spatially distributed streaming data,
evaluating two key performance aspects\fi{} that are essential and supportive for this work: \textbf{numerical stability} and \textbf{wall-clock runtime}.%

\subsection{Numerical Stability Evaluation}
\label{appendix:subsec_opds_ext_1}
\begin{figure}[h]
    \centering
    \includegraphics[width=\figAtwowidth]{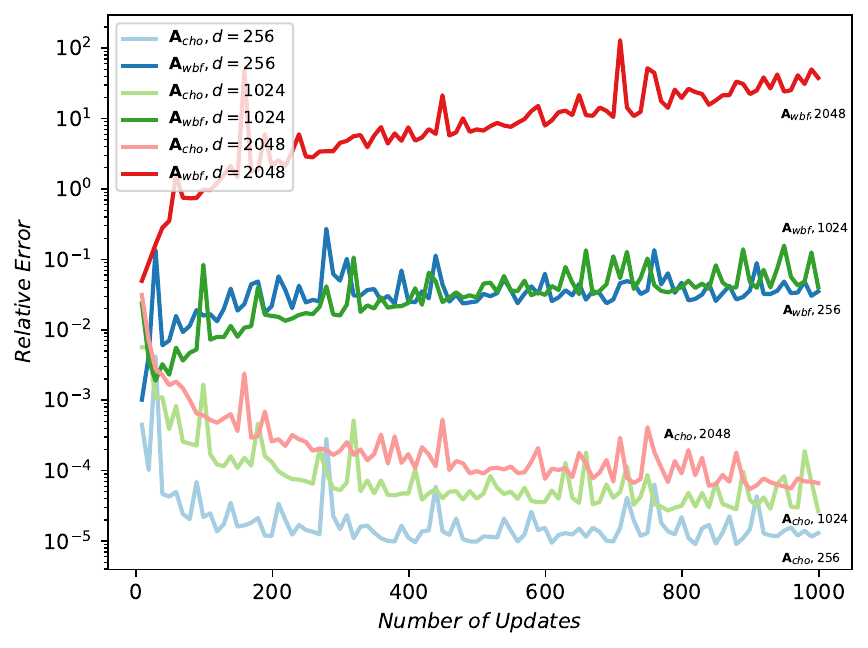}
    \caption{Experiment on numerical stability. The plot shows a comparison between the relative reconstruction error of the Woodbury method and OPDS-FL (Cholesky-driven) approach over 1000 consecutive update iterations, based on three input dimensions.
    The solution maintains consistently low relative reconstruction error rates across all input dimensions. Source:~\ifanon\else our previous work~\fi\citep{roedersampling2024}.}
    \label{fig:appendix_opds_exp_1}
\end{figure}

The reconstruction quality is measured by means of relative error between the directly computed matrix $\mathbf{A}_{\text{dir}}$, updated by a randomized vector $\mathbf{v} \in \mathbb{R}^{d}$ such that $\mathbf{A}_{\text{dir}} \coloneqq (\mathbf{A} + \mathbf{v}\mathbf{v}^\top)^{-1}$,
and the corresponding matrices calculated employing low-rank update strategies based on both Woodbury~\cite{max1950inverting} formula $\mathbf{A}_{\text{wbf}}$ and efficient implementation~\citep{seeger2004} of Cholesky decomposition $\mathbf{A}_{\text{cho}}$ (driving our OPDS-FL component).
Here, $\mathbf{A}_{\text{dir}}$ is the direct dense inverse used only as reference.
The relative error is subsequently defined as
\begin{equation} 
\frac{\norm{\mathbf{A}_{\text{wbf}} - \mathbf{A}_{\text{dir}}}_\text{F}}{\norm{\mathbf{A}_{\text{dir}}}_\text{F}} \ \ \ \text{and} \ \ \ \frac{\norm{\mathbf{A}_{\text{cho}} - \mathbf{A}_{\text{dir}}}_\text{F}}{\norm{\mathbf{A}_{\text{dir}}}_\text{F}} ,
\end{equation}
respectively, with $\norm{\cdot}_{\text{F}}$ employing the Frobenius norm.
The results of this experiment, traced over 1000 iterative updates with input dimensions $d \in \{256, 1024, 2048\}$, are shown in \figurename~\ref{fig:appendix_opds_exp_1}.

\paragraph{Results.} The experiment demonstrates that the approach achieves faster convergence to a stable error and maintains consistently low relative reconstruction error rates throughout the iterative process across all input dimensions, in comparison to the Woodbury update-based method.
It is important to note that, unlike our technique, the Woodbury update-based method becomes unstable at higher input dimensions (see fluctuation for $d$ = 2048), underlining the future-proof suitability of this method for ever-increasing neural network models~\cite{roedersampling2024}. 

\subsection{Wall-clock Runtime Comparison}
\label{appendix:subsec_opds_ext_2}
The authors further evaluate the approach in terms of practical wall clock computation times on a CPU-based system.
Simulating a low-resource edge device without access to GPU acceleration, the exactly same setup is followed as explored in Section~\ref{appendix:subsec_opds_ext_1}.
The computation times are averaged for each dimension over the number of update cycles, see \tablename~\ref{tab:compute}.
\begin{table}[hbt!]\small
\centering
\caption{Algorithmic complexity and average wall-clock runtime comparison considering input dimensions $d = \{256, 1024, 2048\}$. Results calculated over three runs, CPU-only. Source:~\ifanon\else our previous work~\fi\cite{roedersampling2024}.}
\label{tab:compute}
\begin{tabular}{c|ccccccc}
    \toprule
    Method & Complexity & $d$ & \O \ Time\\
    &&& sec $\times 10^{-3}$ \\ %
    \hline
    & & 256 & $0.76 \pm 0.53$ \\
    Direct &  $\mathcal{O}(d^3)$ & 1024 & $16.11 \pm 3.65$ \\
    & & 2048 & $111.86 \pm 14.36$ \\ \hline
    & & 256 & $0.60 \pm 0.50$ \\ 
    Woodbury & $\mathcal{O}(r^2d)$ & 1024 & $10.96 \pm 2.52$ \\
    & & 2048 & $86.55 \pm 13.63$ \\ \hline
    & & 256 & $\textbf{0.60} \bm{\pm} \textbf{0.49}$ \\ 
    OPDS-FL (Cholesky) & $\mathcal{O}(\frac{1}{3} d^3)$ & 1024 &$\textbf{6.68} \bm{\pm} \textbf{1.54}$ \\
    & & 2048 & $\textbf{45.05} \bm{\pm} \textbf{6.47}$ \\
    \bottomrule
\end{tabular}
\end{table}

\paragraph{Results.} Although Woodbury updates (rank $r$=1) have lower theoretical operation counts than Cholesky-based low-rank techniques, they do not consider the nature of the calculations, granting this method a significant advantage in practical implementations:
Cholesky-based updates operate exclusively on lower triangular forms, using cheap back-substitutions and avoiding complex operations for zero elements of the matrix.
This subsequently results in compute times that are on-par or faster in all scenarios~\cite{roedersampling2024}.

\subsection{Diversity Score Visualization}

\label{appendix:subsec_opds_diversity_score}
\begin{figure}[h]
    \centering
    \includegraphics[width=\figAthreewidth]{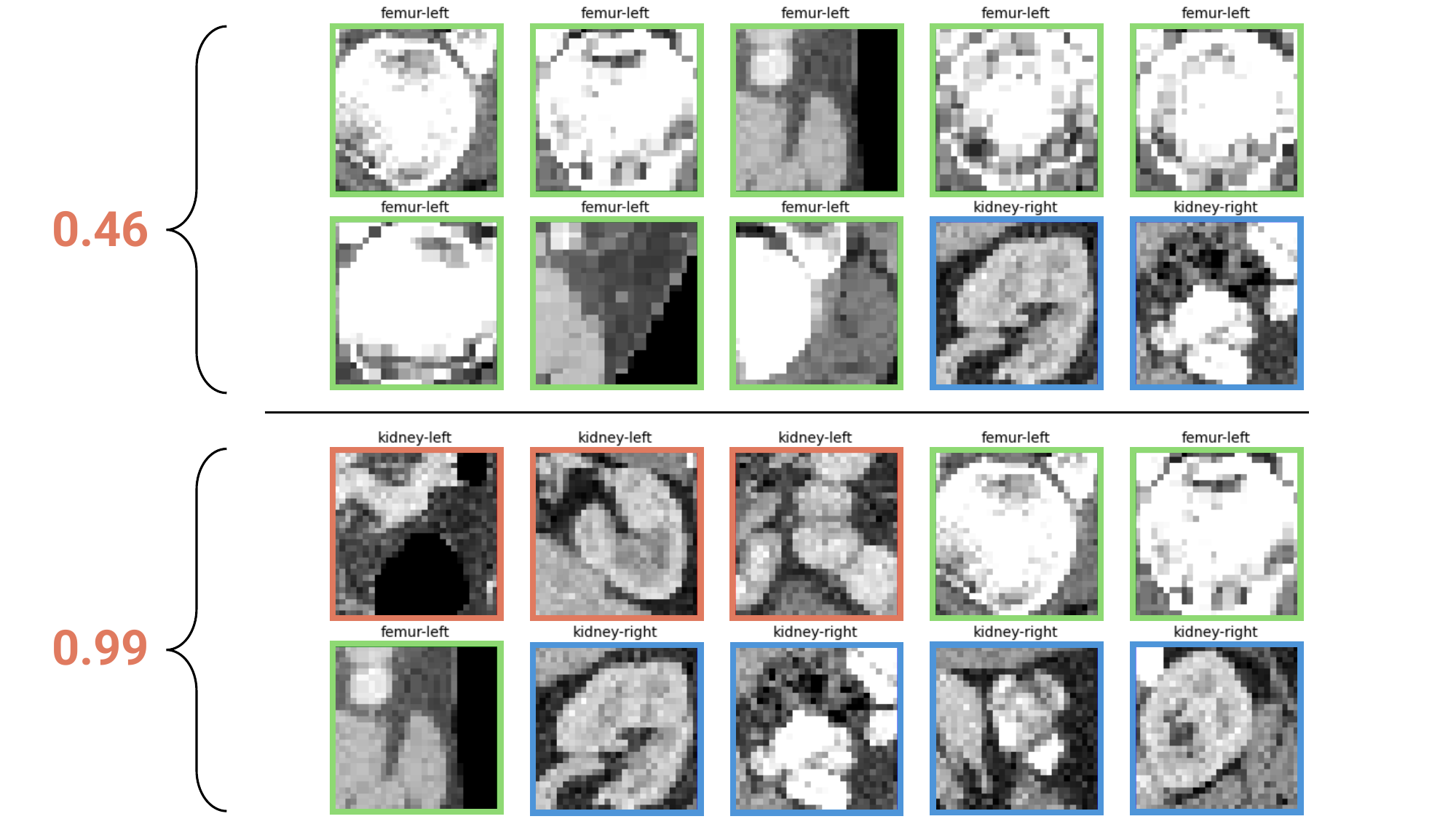}
    \caption{Visualization of two exemplary sampling results with $k_{\mathcal{S}} = 10$, on the medical OrganAMNIST dataset and the occurance of three different classes in the stream.
    The diversity scores (red, detailed in the text) measure the balance of classes in the sampled batch.  Colored image frames indicate class membership.}
    \label{fig:div_score}
    \Description{The figure shows two groups of grayscale medical image patches arranged in a grid, each representing anatomical regions such as the femur and kidney. The upper group, labeled 0.46 in red on the left, contains image patches bordered mostly in green and blue, depicting consistent structures of the femur and kidney with moderate visual variation. The lower group, labeled 0.99, displays a similar grid of patches but with greater visual consistency and sharper focus on the kidney regions. The images in this section are outlined in a mix of red, green, and blue borders, corresponding to different anatomical classes or prediction categories.}
\end{figure}

To illustrate how the diversity score of Equation~\eqref{eq:entropy} unfolds in practice, \figurename~\ref{fig:div_score} shows two exemplary sampling scenarios on the OrganAMNIST dataset with local labeling budget $k_{\mathcal{S}}=10$.

\section{Theoretical Analysis}
\label{app:theory}
This appendix substantiates the coupling claim of Section~\ref{subsec:methodology_stage_2} for drifting streams: the convergence guarantee of the guided updates depends on the sampling stage only through an effective gradient variance, a selection-bias floor, and a bounded sampling-drift term.
All statements analyze the algorithm as specified in Algorithms~\ref{alg:fl_diversity_sampling} and~\ref{alg:fl_guided_update}; drift robustness is validated empirically in Section~\ref{subsec:drift}.

\subsection{Convergence under Drift}
\label{app:theory_main}
The population objective is $F(\theta) = \frac{1}{N}\sum_{i=1}^{N} F_i(\theta)$ with $F_i(\theta) = \mathbb{E}_{(x,y)\sim \mathbb{P}_i}\big[\ell(\theta; x, y)\big]$, where $\mathbb{P}_i$ is the labeled counterpart of the time-averaged stream distribution $\bar{\mathbb{P}}_i$ of Definition~\ref{def:dual_heterogeneity}.
No client optimizes $F_i$ directly: in round $r$, a sampler $\Pi_i$ selects the budgeted subset $D_i$ with $|D_i| \leq k_{\mathcal{S}}$, and training acts on the empirical objective $\hat{F}_i(\theta) = \frac{1}{|D_i|}\sum_{(x,y)\in D_i} \ell(\theta; x, y)$.
Let $\bar{F}^r_i = \mathbb{E}_{\Pi_i}[\hat{F}_i]$ denote the round-$r$ expected sampled objective, $\bar{F}_i = \frac{1}{R}\sum_{r=1}^{R}\bar{F}^r_i$ its horizon average, and $\tilde{F} = \frac{1}{N}\sum_{i=1}^{N} \bar{F}_i$ the surrogate the federation collectively descends; the deviation of $\tilde{F}$ from $F$ is the aggregate selection bias
\begin{equation}\label{eq:bias_floor}
\beta^2 = \sup_{\theta}\, \Big\|\frac{1}{N}\sum_{i=1}^{N}\big(\nabla\bar{F}_i(\theta) - \nabla F_i(\theta)\big)\Big\|^2_2.
\end{equation}
Unweighted informed sampling is biased by design ($\nabla \bar{F}_i \neq \nabla F_i$ in general); stream-uniform sampling is unbiased ($\beta = 0$), and stream-interval sampling is unbiased under exchangeable stream orders, a condition that recurring drift violates (Section~\ref{subsec:drift}).

\begin{assumption}[Smoothness]\label{assump:smooth}
Every per-sample loss $\ell(\cdot\,; x, y)$ is $L$-smooth and bounded below; consequently, so are $F$, each $\bar{F}_i$, and $\tilde{F}$.
\end{assumption}
\begin{assumption}[Two-layer stochastic oracle]\label{assump:oracle}
At every local step of round $r$, client $\mathcal{C}_i$ observes $\mathbf{g}_i(\theta) = \nabla \bar{F}^r_i(\theta) + \zeta$ with $\mathbb{E}[\zeta \mid \theta] = 0$ and $\mathbb{E}\|\zeta\|^2_2 \leq \sigma^2_{\mathrm{eff}} = \sigma^2_{b} + \nu^2(\Pi_i)$: by the law of total variance, the mini-batch noise $\sigma^2_b$ within $D_i$ and the selection variance $\nu^2(\Pi_i) = \mathbb{E}\big\|\nabla\hat{F}_i - \nabla\bar{F}^r_i\big\|^2_2$ add.
The oracle is fresh at every step, ignoring the correlations induced by revisiting $D_i$ across local epochs, as is standard in finite-sample FL analyses.
\end{assumption}
\begin{assumption}[Bounded sampling drift]\label{assump:drift}
There exists $\Delta_{\mathrm{S}} \geq 0$ such that, for all $i$ and $r$, $\sup_\theta \big\|\nabla\bar{F}^r_i - \nabla\bar{F}_i\big\|_2 \leq \Delta_{\mathrm{S}}$.
\end{assumption}

\begin{theorem}[Sampler-dependent convergence under drift]\label{thm:sampled_scaffold}
Let Assumptions~\ref{assump:smooth}--\ref{assump:drift} hold, let $\tilde{F}_0 = \tilde{F}(\theta^0_{\mathcal{S}}) - \inf_\theta \tilde{F}(\theta)$, and run \textsc{STORMGUIDE-FL} for $R$ rounds of $K$ guided steps on $|N^{\text{sub}}|$ participating clients per round with the step sizes of \tcite[Karimireddy et al.]{pmlr-v119-karimireddy20a}. Then, at bound level,
\begin{equation}\label{eq:main_rate}
\min_{r \leq R}\, \mathbb{E}\big\|\nabla F(\theta^r_{\mathcal{S}})\big\|^2_2 \;\leq\; 2\,\varepsilon_R\big(\sigma_{\mathrm{eff}}\big) + 2\beta^2 + \mathcal{O}\big(\Delta^2_{\mathrm{S}}\big),
\end{equation}
where
\begin{equation*}
\varepsilon_R\big(\sigma_{\mathrm{eff}}\big) = \mathcal{O}\!\left(\frac{\sigma_{\mathrm{eff}}\sqrt{L \tilde{F}_0}}{\sqrt{|N^{\text{sub}}|\, K R}}\right) + c_R
\end{equation*}
is the non-convex convergence rate of \textsc{SCAFFOLD} applied to $\tilde{F}$ and $c_R \geq 0$ collects higher-order terms that depend on the sampler only through $\tilde{F}_0$.
In the stationary regime ($\Delta_{\mathrm{S}} = 0$, all $\bar{F}^r_i = \bar{F}_i$) the bound holds with the $\mathcal{O}(\Delta^2_{\mathrm{S}})$ term absent and no bound-level step is required.
\end{theorem}
\begin{proof}[\proofsketch]
Each guided step (Equation~\eqref{eq:guided_local_update}) applies $\theta_i \leftarrow \theta_i - \eta_l\big(\mathbf{g}_i(\theta_i) - \mathbf{c}_i + \mathbf{c}_{\mathcal{S}}\big)$, and the variate and server updates (Equations~\eqref{eq:client_variate_update},~\eqref{eq:server_aggregate},~\eqref{eq:server_update}) are exactly those of \textsc{SCAFFOLD}~\citep{pmlr-v119-karimireddy20a}.
By Assumptions~\ref{assump:oracle} and~\ref{assump:drift}, $\mathbf{g}_i$ is an oracle for the round-independent $\nabla\bar{F}_i$ with deterministic bias at most $\Delta_{\mathrm{S}}$ and variance at most $\sigma^2_{\mathrm{eff}}$.
For $\Delta_{\mathrm{S}} = 0$ the non-convex \textsc{SCAFFOLD} guarantee applies verbatim to $\tilde{F}$, yielding $\min_{r} \mathbb{E}\|\nabla\tilde{F}\|^2_2 \leq \varepsilon_R(\sigma_{\mathrm{eff}})$ without any bound on gradient dissimilarity across clients; for $\Delta_{\mathrm{S}} > 0$, absorbing the bias cross term via Young's inequality degrades the descent coefficient by a constant factor and adds an $\mathcal{O}(\Delta^2_{\mathrm{S}})$ residual, as for SGD with biased oracles~\citep{ajalloeian2020biased}.
Finally, $\nabla F = \nabla\tilde{F} - \frac{1}{N}\sum_i\big(\nabla\bar{F}_i - \nabla F_i\big)$, so $\|\nabla F\|^2_2 \leq 2\|\nabla\tilde{F}\|^2_2 + 2\beta^2$ by Young's inequality and Equation~\eqref{eq:bias_floor}.
\end{proof}
\begin{remark}[Drift enters through Definition~\ref{def:dual_heterogeneity}]\label{rem:drift_link}
For order-oblivious samplers such as stream-uniform sampling, $\bar{F}^r_i$ is the loss under the round-$r$ stream mixture; if per-sample gradients are bounded by $G$ (as in Proposition~\ref{prop:dp}) and $d$ is total variation, joint convexity of $d$ yields $\Delta_{\mathrm{S}} \leq 2G\, \Delta_{\mathrm{time}}$, so the intra-client drift condition of Definition~\ref{def:dual_heterogeneity} enters the guarantee directly.
For adaptive samplers, $\Delta_{\mathrm{S}}$ additionally reflects sensitivity to stream order: the order-invariant coverage objective of OPDS-FL keeps the selected cross-section stable across rounds, which Section~\ref{subsec:drift} confirms empirically, whereas interval sampling keeps $\Delta_{\mathrm{S}}$ small but pays under recurring drift through the aliasing-induced bias $\beta$.
\end{remark}

\subsection{Sampler-Dependent Constants}
\label{app:theory_opds}
Let $\mathbf{\Sigma}_i = \mathbb{E}\big[f_i(x) f_i(x)^\top\big]$ denote the population feature second moment of stream $\mathcal{U}_i$, $d_{\mathrm{eff}} = \operatorname{tr}\big(\mathbf{\Sigma}_i(\mathbf{\Sigma}_i + \lambda \mathbf{I})^{-1}\big) \leq d$ its effective dimension at the regularization level $\lambda$ of Algorithm~\ref{alg:fl_diversity_sampling}, and $\chi_i \geq 1$ the leverage coherence of the stream (maximal over average ridge leverage score), which grows under redundant or skewed feature distributions.

\begin{lemma}[OPDS-FL is budget-paced streaming ridge-leverage sampling]\label{lem:leverage}
Let $\hat{\mathbf{\Sigma}}^t_i$ be the regularized second moment of the accepted set maintained by Algorithm~\ref{alg:fl_diversity_sampling} (initialized at $\lambda\mathbf{I}$), and let $\tau^t_\lambda(x) = f_i(x)^\top (\hat{\mathbf{\Sigma}}^t_i)^{-1} f_i(x)$ denote the ridge leverage score of $x$ with respect to the accepted set. Then the acceptance probability of Equation~\eqref{eq:prob_update} satisfies
\begin{equation*}
p^t_i \;=\; q^t_i \cdot \frac{\tau^t_\lambda(x^t_i)}{\bar{\tau}^t}, \qquad \bar{\tau}^t = \frac{1}{t}\sum_{j=1}^{t} \tau^t_\lambda(x^j_i),
\end{equation*}
immediately from Equations~\eqref{eq:covariance_tracking} and~\eqref{eq:prob_update}.
\end{lemma}

\begin{proposition}[Variance and bias of OPDS-FL]\label{prop:variance_bias}
Assume per-sample gradients are head-dominated, $\nabla_\theta\, \ell(\theta; x, y) \approx \mathbf{M}(\theta, y)\, f_i(x)$ with $\sup \|\mathbf{M}\|_{\mathrm{op}} \leq \kappa$. Then:
(i) uniform sampling has selection variance $\nu^2_{\mathrm{unif}} = \mathcal{O}\big(\kappa^2 \operatorname{tr}(\mathbf{\Sigma}_i)/k_{\mathcal{S}}\big)$, and determinantal repulsion reduces it further under redundancy~\citep{zhang2017dppminibatch};
(ii) leverage-proportional sampling covers every curvature-bearing feature direction (an importance-weighted spectral $(1\pm\epsilon)$-approximation of $\mathbf{\Sigma}_i$, up to budget pacing and probability clipping) at budget $k_{\mathcal{S}} = \tilde{\Omega}(d_{\mathrm{eff}})$ independently of the coherence $\chi_i$, whereas uniform sampling requires the budget inflated by order $\chi_i$~\citep{cohen2016online, cohen2017ridge, drineas2016randnla};
(iii) the price is the bias $b_i = \nabla\bar{F}_i - \nabla F_i$ induced by training unweighted on the leverage-tilted selection, with $\|b_i\|_2 \leq \kappa\, \|\mu^{\Pi}_i - \mu_i\|_2$ for the tilted and true feature-residual means and $\beta^2 \leq \max_i \|b_i\|^2_2$; $b_i$ vanishes if the leverage scores are uniform.
\end{proposition}
\noindent OPDS-FL therefore improves the guarantee of Theorem~\ref{thm:sampled_scaffold} over an unbiased baseline when its variance saving outweighs its bias floor, predicting the largest gains on redundant, strongly non-IID streams (large $\chi_i$), consistent with Section~\ref{subsec:exp_2}; we do not claim universal dominance.
Sampling and guidance also compound at bound level: \textsc{FedAvg} guarantees carry an unavoidable cross-client heterogeneity term that is independent of the within-client sampler~\citep{pmlr-v119-karimireddy20a}, while Equation~\eqref{eq:main_rate} carries none, so reducing $\nu^2(\Pi_i)$ pays off disproportionately under guided updates: the super-additive interaction measured in Section~\ref{subsec:exp_2}.
Finally, by construction, $|D_i| \leq k_{\mathcal{S}}$ holds deterministically ($q^t_i = 0$ once the budget is spent), each observation costs $\mathcal{O}(d^2)$ time and memory via the Cholesky update of Equation~\eqref{eq:chol_update}, and the per-round communication is identical to \textsc{SCAFFOLD}.

\section{Privacy Analysis}
\label{app:privacy}
\textsc{STORMGUIDE-FL} follows the standard FL privacy posture: raw stream observations, assigned labels, and all sampling decisions remain on the client; only the deltas $(\Delta\theta_i, \Delta\mathbf{c}_i)$ are transmitted.
Against an honest-but-curious server, the released updates admit a standard $\varepsilon$-differential-privacy (DP) extension~\cite{dwork2014dp}: two datasets $D, D'$ are \emph{adjacent} if they differ in a single sample-label pair, and a randomized mechanism $\mathcal{M}$ is $\varepsilon$-DP if, for all adjacent $D, D'$ and measurable $S$, $\Pr[\mathcal{M}(D) \in S] \leq e^{\varepsilon} \Pr[\mathcal{M}(D') \in S]$.
For the analysis, the client performs $K$ full-batch gradient steps with learning rate $\eta_l$ on its local training set $D_i$, and $d_\theta$ denotes the number of model parameters.

\begin{proposition}[$\varepsilon$-DP of \textsc{STORMGUIDE-FL} updates]
\label{prop:dp}
Assume each per-sample loss is $L$-smooth (Assumption~\ref{assump:smooth}) with gradient norm uniformly bounded by $G$, i.e., $\|\nabla_{\theta}\, \ell(g_i(f_i(x; \theta)), y)\|_2 \leq G$ for all clients $i$ and samples $(x, y)$, and let $\eta_l \leq 1/(LK)$. Let the client release
$\widetilde{\Delta\theta}_i = \Delta\theta_i + \zeta$ with $\zeta \sim \mathrm{Lap}(\lambda_\varepsilon)^{d_\theta}$ applied coordinate-wise, where
\begin{equation*}
\lambda_\varepsilon \;=\; \frac{2(e-1)\,\eta_l K G \sqrt{d_\theta}}{\varepsilon\,|D_i|}.
\end{equation*}
Then the release of $\widetilde{\Delta\theta}_i$ satisfies $\varepsilon$-DP with respect to $D_i$, and the remote guidance variate delta computed from the noised quantity, $\widetilde{\Delta\mathbf{c}}_i = -\mathbf{c}_{\mathcal{S}} - \frac{1}{K\eta_l}\widetilde{\Delta\theta}_i$, inherits the same guarantee.
\end{proposition}
\begin{proof}[\proofsketch]
Let $\delta_k$ denote the parameter distance after $k$ steps between the trajectories trained on adjacent sets $D_i, D'_i$.
Replacing one sample changes the full-batch gradient at any fixed $\theta$ by at most $2G/|D_i|$ in $\ell_2$-norm, and $L$-smoothness bounds the additional deviation between the trajectories by $L\delta_k$, yielding $\delta_{k+1} \leq (1 + \eta_l L)\,\delta_k + 2\eta_l G/|D_i|$ with $\delta_0 = 0$.
Unrolling gives $\delta_K \leq \frac{2G}{L\,|D_i|}\big((1+\eta_l L)^K - 1\big) \leq 2(e-1)\,\eta_l K G/|D_i| =: \Delta_2$ for $\eta_l \leq 1/(LK)$, so the $\ell_1$-sensitivity of $\Delta\theta_i$ satisfies $\Delta_1 \leq \sqrt{d_\theta}\,\Delta_2$.
The Laplace mechanism with scale $\lambda_\varepsilon = \Delta_1/\varepsilon$ is $\varepsilon$-DP~\cite{dwork2014dp}.
$\widetilde{\Delta\mathbf{c}}_i$ is a deterministic function of $\widetilde{\Delta\theta}_i$ and quantities independent of the current round's data (cf.\ Equation~\eqref{eq:client_variate_update}), so the guarantee carries over by post-processing immunity.
\end{proof}

Participation in $T$ rounds composes to $(T\varepsilon)$-DP by basic composition, provided the client also persists its variate from the noised release, $\mathbf{c}_i \gets \mathbf{c}_i + \widetilde{\Delta\mathbf{c}}_i$, so that later rounds access $D_i$ only through its $\varepsilon$-DP output (the local model carries no such state, as it is re-initialized from $\theta_{\mathcal{S}}$ every round); tighter accounting via the Gaussian mechanism and moments accountant~\cite{abadi2016dpsgd} applies unchanged, since the mechanism is output perturbation of a bounded-sensitivity release.
Notably, the sampling stage (Algorithm~\ref{alg:fl_diversity_sampling}) operates exclusively on locally computed feature statistics and consumes no privacy budget, and the added noise perturbs the aggregated server remote guidance variate $\mathbf{c}_{\mathcal{S}}$ only by a zero-mean term of variance $O(\lambda_\varepsilon^2 d_\theta / |N^{\text{sub}}|)$: the guidance correction degrades gracefully, reducing in the uninformative limit to the OPDS-FL-only setting evaluated in Section~\ref{subsec:exp_2}.

\end{document}